\documentclass[12pt]{article}

\usepackage{amsmath,amsthm, amsfonts, amssymb, amsxtra, amsopn}
\usepackage{pgfplots}
\usepgfplotslibrary{colorbrewer}
\pgfplotsset{compat = 1.15, 
			 cycle list/Set1-3} 
\usetikzlibrary{pgfplots.statistics, pgfplots.colorbrewer} 
\usepackage{pgfplotstable}
\usepackage{graphicx,grffile}
\usepackage{multirow}
\usepackage{booktabs}
\usepackage{tcolorbox}
\usepackage{algorithm} 
\usepackage[noend]{algpseudocode} 
\usepackage{listings}
\usepackage{cmap}
\usepackage{colortbl}
\usepackage{adjustbox}
\usepackage{epsfig}

\usepackage[tableposition=top,font=small,skip=5pt]{caption}
\usepackage{subcaption}
\usepackage{makecell}

\usepackage[explicit]{titlesec}

\usetikzlibrary{patterns}

\def\bbb#1{{\color{blue}#1}}
\def\com#1{\bbb{\texttt{/\kern-1.5pt /} #1}}

\def\tau{\mathcal{T}}

\PassOptionsToPackage{hyphens}{url}
\PassOptionsToPackage{table}{xcolor}

\usepackage{hyperref}
\hypersetup{colorlinks=true,linkcolor=black,citecolor=black,urlcolor=blue,filecolor=black}
\hypersetup{pdfpagemode=UseNone,pdfstartview=}

\definecolor{darkgreen}{rgb}{0.125,0.5,0.169}

\usepackage[shortlabels]{enumitem}
\setlist[itemize]{noitemsep, topsep=0pt}

\advance\oddsidemargin by -0.45in
\advance\textwidth by 0.9in

\advance\topmargin by -0.5in
\advance\textheight by 1.0in

\def\eref#1{(\ref{#1})}

\long\def\symbolfootnotetext[#1]#2{\begingroup%
  \def\thefootnote{\fnsymbol{footnote}}\footnotetext[#1]{#2}\endgroup}

\newcommand\dunderline[3][-1pt]{{%
      \sbox0{#3}%
      \ooalign{\copy0\cr\rule[\dimexpr#1-#2\relax]{\wd0}{#2}}}}
\def\uuu{\kern-1pt\dunderline{0.75pt}{\phantom{M}}}

\def\matches#1{{\cal #1}}
\DeclareMathOperator{\thth}{th}
\DeclareMathOperator{\out}{outliers}
\DeclareMathOperator{\inli}{inliers}
\DeclareMathOperator{\drift}{drift}
\DeclareMathOperator{\ddrift}{drift}
\DeclareMathOperator{\control}{control}

\DeclareMathOperator{\test}{test}
\DeclareMathOperator{\train}{train}
\DeclareMathOperator{\periodic}{periodic}
\DeclareMathOperator{\static}{static}
\DeclareMathOperator{\svm}{SVM}
\DeclareMathOperator{\mlp}{MLP}
\DeclareMathOperator{\rf}{RF}
\DeclareMathOperator{\xgb}{XGB}

\DeclareMathOperator{\mmd}{MMD}

\DeclareMathOperator{\hilbertSpace}{\mathcal{H}}
\def\pvalue{$p$-\hbox{value}}

\DeclareMathOperator{\oc}{\textsc{ocsvm}}
\DeclareMathOperator{\mk}{\textsc{mk-m}eans}
\DeclareMathOperator{\mmds}{\textsc{mmd}}
\DeclareMathOperator{\clusters}{\texttt{clusters}}
\DeclareMathOperator{\kernel}{\tt kernel}

\DeclareMathOperator{\solver}{\tt solver}

\DeclareMathOperator{\fame}{\ast}
\DeclareMathOperator{\nestimators}{\tt n\_estimators}
\DeclareMathOperator{\nunits}{\tt n\_units}
\DeclareMathOperator{\activation}{\tt activation}

\DeclareMathOperator{\relu}{relu}
\DeclareMathOperator{\logistic}{logistic}
\DeclareMathOperator{\adam}{adam}
\DeclareMathOperator{\sgd}{sgd}
\DeclareMathOperator{\mxD}{\tt max\_depth}
\DeclareMathOperator{\learningrate}{\tt learning\_rate}
\DeclareMathOperator{\reference}{ref}
\DeclareMathOperator{\new}{new}
\DeclareMathOperator*{\argmin}{arg\,min}

\DeclareMathOperator*{\BCE}{BCE}

\def\Agent{\texttt{Agent}}
\def\Airpush{\texttt{Airpush}}
\def\Boxer{\texttt{Boxer}}
\def\Malap{\texttt{Malap}}
\def\SMSreg{\texttt{SMSreg}}

\def\zz{\phantom{0}}

\title{Concept Drift Detection and Adaptive Retraining of Malware Classification Models}

\author{Christofer Washington Berruz Chungata\footnotemark[1]\ \ \ 
Martin  Jure\v{c}ek\footnotemark[2]\ \ \ \\
Katerina Potika\footnotemark[1]\ \ \ 
William B. Andreopoulos\footnotemark[1]\ \ \ 
Mark Stamp\footnotemark[1]\,\,\footnotemark[3]} 

\begin{document}

\symbolfootnotetext[1]{Department of Computer Science, San Jose State University}
\symbolfootnotetext[2]{Faculty of Information Technology, Czech Technical University in Prague}
\symbolfootnotetext[3]{mark.stamp$@$sjsu.edu}

\maketitle

\abstract
Concept drift refers to changes over time in the 
statistical properties of data, as compared to the data 
that was used to train a learning model.
Machine learning models for malware detection or classification
are particularly susceptible to performance
degradation caused by concept drift, as
attackers constantly modify existing malware.
In this chapter, we analyze two machine learning-based 
approaches to automated
concept drift detection---a novel approach based on 
One-Class Support Vector Machines (OCSVM)
and a previously-studied technique
based on Minibatch $K$-Means (MK-Means). For
comparison we also consider  
Maximum Mean Discrepancy (MMD), 
a statistical technique for detecting changes in multidimensional data. 
We conduct an extensive series of experiments 
comparing the effectiveness of four learning 
models, namely, Multilayer Perceptron (MLP),
Random Forest (RF), Support Vector Machines (SVM),
and eXtreme Gradient Boosting (XGB).
For each of these models, we consider three distinct scenarios: 
A static scenario where no model retraining occurs, 
a periodic scenario where models are constantly
retrained irrespective of concept drift,
and a drift-aware scenario where
models are only retrained when concept drift is detected.
Under the drift-aware scenario, 
we analyze the tradeoff between accuracy and training efficiency
using Pareto Front analysis.
We find that all three concept drift detection techniques 
achieve classification accuracy comparable to
periodic retraining, while offering substantially greater efficiency in terms
of the number of models that must be retrained.
In addition, drift-aware retraining based on
our OCSVM technique generally outperforms 
the MK-Means and MMD approaches.
Overall, these results provide strong evidence that we are able to accurately 
detect concept drift in malware classification models. 
Furthermore, our concept drift detection techniques
are efficient and practical, and the process of updating 
learning models can easily be fully automated.

\bigskip


\section{Introduction}

Change over time in the statistical properties of data on which learning models 
have been trained is known as concept drift.
Given that machine learning models learn from data, concept drift can 
degrade the performance of such models during inference.

Concept drift is a particularly important problem in malware detection as malware is constantly evolving.
According to~\cite{aycock, stamp2023malware}, attackers modify malware for two purposes---to add 
or modify functionality and to evade detection. As a result, developers and researchers using 
machine learning models for malware detection or classification must take concept drift into consideration 
when deploying these models in real-world systems.

The number of cyberattacks due to malware, such as ransomware attacks, 
has increased significantly in recent years~\cite{bertia2022ransomware}.
A single malware sample that is incorrectly classified as benign can wreak 
havoc on a system by bringing down machines or holding data hostage for ransom.

One solution to mitigate the effects of concept drift is to constantly retrain the models 
used for detection. However, the resource demands of this approach may be
substantial, in terms of computational resources and energy usage.
In an era where significant research efforts focus on minimizing the resource consumption 
of complex machine learning models, such as Deep Neural Networks (DNN) and 
Large-Language Models (LLM)~\cite{energyTrendsInAI,leo}, constant retraining 
would seem to be a step backwards.

In this research, we conduct experiments involving two machine learning-based 
approaches for concept drift detection in malware classification models---one based on 
Minibatch $K$-Means (MK-Means) and another 
based on One-Class Support Vector Machines (OCSVM). For comparison, we also consider 
concept drift detection using the well-known statistical technique of 
Maximum Mean Discrepancy (MMD), 
which can be applied to detect changes in multidimensional data.
To demonstrate the effectiveness of these concept drift detection approaches, 
we consider the following three scenarios, which are discussed 
in more detail in Section~\ref{sec:setup}.

\begin{enumerate}[i)]
    \item A classification model is trained on the initial temporal segment of the training data, 
    and no retraining occurs. This represents a scenario where concept drift is not considered.
    We refer to this as the \textit{static} scenario.
    \item As in~i), a classification model is trained on the initial temporal segment of the training data.
    Then the model is retrained at regular periodic intervals. This represents a scenario where 
    concept drift is considered, but no effort is made to minimize the number of 
    models that must be retrained. We refer to this as the \textit{periodic retraining} scenario.
    \item As in~ii), a classification model is trained on the initial temporal segment of the training data 
    and concept drift is considered. However, instead of periodic retraining, we only retrain when 
    concept drift is detected via one of the methods mentioned above. This represents a scenario 
    where we account for concept drift, but we also attempt to minimize model retraining. 
    We refer to this as the \textit{drift-aware retraining} scenario.
\end{enumerate}

For each of these three scenarios, we consider four classification models, namely, 
Multilayer Perceptron (MLP), Random Forest (RF), eXtreme Gradient Boosting (XGB), 
and Support Vector Machines (SVM). Furthermore, for each of these model-scenario 
combinations, we consider~20 distinct combinations of malware families.
Thus, we conduct~80 static experiments, 80 periodic retraining experiments, 
and~240 drift-aware retraining experiments (80 using each of the OCSVM, 
MK-Means, and MMD drift detectors), giving us a total of~400 distinct experiments.

The remainder of this chapter is organized as follows.
Section~\ref{chap:background} provides relevant background information on concept drift, 
MK-Means, OCSVM, MMD, and the learning models used in our classification experiments.
In Section~\ref{chap:implementation}, we describe the dataset used in this research, 
and the development environment for our experiments. In addition, this section describes 
the static, periodic retraining, and drift-aware retraining scenarios in detail.

Section~\ref{chap:results} presents results and analysis for all of our experiments.
We demonstrate that using drift-aware retraining yields a large increase in accuracy, 
as compared to the static case, and only a slight decrease in accuracy, as compared 
to periodic retraining. We further analyze the drift-aware retraining case, and provide 
detailed information on the tradeoff between accuracy and efficiency.
We conclude the chapter in Section~\ref{chap:conclusion}, 
where we also provide suggestions for future work.

Note that this chapter expands on and extends the research that appeared in the paper~\cite{mmd}.
Specifically, the MMD technique presented in this chapter is analyzed in~\cite{mmd}, which
in the present chapter, merely serves as a baseline for comparison of the MK-Means and 
OCSVM techniques. 

\section{Background}\label{chap:background}

In this section, we first discuss concept drift and malware detection.
Then we provide details on the OCSVM, MK-Means, and MMD techniques, 
which form the basis for our automated concept drift detection experiments.
Finally, we briefly introduce the learning models that are used in our experiments.

\subsection{Concept Drift}

Concept drift is defined as changes in the statistical properties of data over time, 
which can affect the performance of machine learning models~\cite{driftSurvey2022}.
Machine learning models deployed in real-world systems may be adversely affected by concept drift, 
as the data used for training the model can age out or become irrelevant.
This is certainly the case for machine learning-based malware detection systems, 
as malware families generally tend to evolve over time.

Concept drift can occur abruptly, gradually, or it can be recurring~\cite{driftSurvey2022}.
Abrupt concept drift is the easiest to detect, while gradual and recurring concept drift require 
more sophisticated methods that incorporate previous knowledge of the data and the model~\cite{driftPerformanceBernoulliMapToshio, driftPerformanceWeightedMovingAverageRoss,
driftPerformanceFuzzyDrift, driftPerformanceAdaptiveOnlineZhang}.

According to~\cite{driftSurvey2022}, any concept drift detection technique 
can be classified into one of four categories, namely, data distribution-based, 
performance-based, hybrid, and context-based.
Performance-based methods are considered more reliable than data distribution-based methods, 
as they focus on instances where the model is performing poorly~\cite{driftSurvey2022}.
However, such methods rely on the ability to measure well-defined performance metrics 
(e.g., accuracy, precision, recall, F1-score), and when using machine learning models to 
classify malware, such information is generally lacking.

Given the limitations of performance-based methods in the malware domain, 
we focus our attention on data distribution-based methods. That is, we require drift detection 
techniques that can be performed by directly analyzing the data, without relying on 
measures of model performance. We validate that our drift detection approaches succeed 
by comparing the performance of classification models under three distinct 
scenarios---details on these scenarios are given in Section~\ref{sec:setup}.

\subsection{Malware Detection}

Malware detection is the process of identifying whether a given piece of software is 
malicious (malware) or benign. Malware detection is a vitally important area of research 
for at least three reasons. First, false negatives (malware misclassified as benign) can lead 
to significant damage to a system, such as data loss and system downtime.
Second, false positives (benign misclassified as malware) can lead to significant 
inconvenience to users. Even a relatively small false positive rate can result in 
a ``boy who cried wolf'' syndrome, where users lose trust in the system and ignore 
warnings. Third, malware is constantly evolving, making it challenging to detect new 
malware variants~\cite{aycock, stamp2023malware}.

Over the past~20 years, malware research has shifted dramatically, to where the focus today 
is dominated by machine learning (ML) and deep learning (DL) techniques.
From classical ML models, such as Support Vector Machines (SVM) and 
Random Forests (RF), to DL models, such as Convolutional Neural Networks (CNN) 
and Recurrent Neural Networks (RNN), learning models are now routinely applied to 
the malware detection and classification problems~\cite{malwareDetectionSurveyAzeem2024,
malwareDetectionSurveyMerabet2019, malwareDetectionSurveyRedhu2024}.

In recent years, the impact of concept drift on malware detection has gained some attention 
from researchers---although certainly not as much attention as the importance
of the topic warrants. For example, the authors of~\cite{li2025conceptdriftmalwareDetection} 
proposed a novel technique for detecting and classifying concept drift in malware by learning 
supposedly drift-invariant features from control flow graphs. Another example of such research 
can be found in~\cite{alam2024morphMalwareDetection}, which introduced MORPH, 
a pseudo-label-based concept drift adaptation method for neural networks as used for malware detection.
Additionally, the paper~\cite{he2024dreamMalwareDetection} considers DREAM, 
a semi-supervised system that enhances drift detection in deep learning-based malware 
classifiers. Note that the approaches considered in our research 
differ from~\cite{alam2024morphMalwareDetection, he2024dreamMalwareDetection, 
li2025conceptdriftmalwareDetection}, as we use unsupervised machine learning techniques 
that are faster, simpler, and more efficient to implement.
Our methodology builds on and extends the work in~\cite{mishra2024cluster}.

We note in passing that research into malware evolution is related to concept drift, 
in the sense that such research deals with changes to malware over time.
However, in concept drift research, the focus is on how these changes affect 
the performance of machine learning models, while malware evolution research 
deals with the changes themselves. Hidden Markov Models (HMM)~\cite{tupadha2022mlmalwareevolution},
Word2Vec~\cite{paul2020wordembedding}, SVMs~\cite{wadkar2020malwareevolution}, 
and hierarchical tree models~\cite{zhang2022manage} have all been used to 
study malware evolution.

Research into concept drift detection for malware requires data that contains 
not only a significant number of malware samples over an extended period of time, 
but also include reliable timestamps. This is challenging because the time when a 
malware sample was originally created is not generally available. Therefore, modification 
or collection dates are used as proxies for the creation date. Prior to the KronoDroid 
dataset~\cite{guerra2021kronodroid}, other malware datasets used for research involving 
temporal aspects of malware analysis included the Malicia dataset~\cite{nappa2014malicia} 
and the malware collection in~\cite{kim2018peheader}.

\subsection{Concept Drift Detection Techniques}\label{sect:driftTechniques}

In this section, we introduce the two unsupervised learning techniques that we employ 
for concept drift detection, namely, Minibatch $K$-Means (MK-Means)
and One-Class Support Vector Machines (OCSVM).
Furthermore, we introduce Maximum Mean Discrepancy (MMD), an established technique 
to detect data drift in multidimensional data. Additional details on precisely how we employ these 
techniques for concept drift detection are provided in Section~\ref{sec:setup}.

\subsubsection{Minibatch $K$-Means}\label{sec:mkmeans_intro}

Minibatch $K$-Means (MK-Means)~\cite{alonso2013kmeansvsminibatch} is a more 
scalable variant of the well-known $K$-Means clustering algorithm~\cite{originalKMeans}.
While many $K$-means variants exist, such as 
$K$-Means\texttt{++}~\cite{kmeansComprehensiveReview2023}, MK-Means has been 
optimized for large datasets by using stochastic optimization principles. The main difference 
between $K$-Means and \hbox{MK-Means} is that the latter uses small random subsets of the 
data (i.e., minibatches) to update the centroids, instead of using the entire dataset at each iteration.

As with $K$-means, MK-Means is an unsupervised clustering algorithm that uses a hill climb 
approach to find improved centroids of clusters. Formally, the \hbox{$K$-Means} algorithm aims to 
minimize Within-Cluster Sum of Squares (WCSS), which is defined as
$$
    \displaystyle\argmin_{\{S_1,\ldots,S_K\}} \sum_{i=1}^K \sum_{x \in S_i} \| x - \mu_i \|^2
$$
where~$\mu_i$ is the mean, or centroid, of the data points in cluster~$S_i$, 
and the nonempty subsets~$S_1,S_2,\ldots,S_K$ of~$S$ form a partition 
of the set~$S$. Note that a centroid~$\mu_i$ need not be an actual data point 
in its cluster~$S_i$.

Given that it is a hill climb algorithm, MK-Means is sensitive to the initial placement of the centroids.
Hence, MK-Means can produce different clustering results, even for a fixed number of clusters~$K$.

Intuitively, a good clustering result should produce clusters that are well separated from each other, 
while any individual cluster should be relatively compact. The \textit{silhouette coefficient} provides 
a single value that incorporates both of these desirable aspects of cluster quality.

For a given clustering, of the~$n$ data points~$\{x_1,x_2,\ldots,x_n\}$, the silhouette coefficient 
of a specific data point~$x_i$ is defined as
$$
    s(i) = \frac{b(i) - a(i)}{\max\big(a(i), b(i)\big)}
$$
where~$a(i)$ is the average distance between~$x_i$ and all other points in its same cluster, 
and~$b(i)$ is the minimum of the average distances from~$x_i$ to points in each of the other clusters.
For a reasonable clustering, we expect~$b(i) > a(i)$, in which case
$$
    s(i) = 1 - \frac{a(i)}{b(i)}
$$
In this form, we see that a silhouette coefficient value~$s(i)$ closer to~1 
represents a better clustering result for~$x_i$ than values closer to~0.
Thus, the larger the average silhouette coefficient
$$
    s = \frac{1}{n} \sum_{i=1}^n s(i)
$$
the better the overall quality of a given clustering.

For concept drift detection, we use MK-Means to cluster consecutive (overlapping) 
pairs of temporal batches of samples, and we compute the average silhouette coefficient 
for each such clustering. If the change in average silhouette coefficient from one pair of 
batches to the next exceeds a specified threshold, this implies that the features representing 
the underlying samples have changed, and hence concept drift has occurred.
More details on MK-Means for concept drift detection are provided in Section~\ref{sect:detect}.
We note that our use of MK-Means for concept drift detection in malware closely follows the 
approach in~\cite{mishra2024cluster}.

\subsubsection{One-Class Support Vector Machines}\label{sec:ocsvm}

Support Vector Machines (SVM) are supervised machine learning models used for classification 
and regression tasks. One-Class SVMs (OCSVM) are a variant of SVMs that are trained in an 
unsupervised mode, and hence labeled data is not required. OCSVMs were originally 
introduced in~\cite{scholkopf2001ocsvm}.

OCSVMs are used widely for outlier detection~\cite{ocsvmdreiseitl2010,ocsvmDocumentClassification}.
Outlier detection is a generic problem where, for a given dataset, there exists two regions
defined as inliers and outliers. The outliers differ significantly from the inliers, 
while the inliers form the bulk of the data.
Many techniques exist for outlier detection, including Robust Covariance~\cite{robustCovariance}, 
Isolation Forest~\cite{isolationForest}, and Local Outlier Factor~\cite{localOutlierFactor}; 
refer to~\cite{scikitLearnOutlierDetection} for a visual comparison of the results of these 
techniques on a synthetic dataset.

OCSVM tries to find the region where the training samples are representative of the 
dominant distribution of the data. OCSVMs include a regularization parameter~$\nu \in (0, 1)$, 
which sets an upper limit on the fraction of training data points that can be considered outliers.
Larger values of~$\nu$ can lead to a more complex, but looser, decision boundary, 
while smaller values of~$\nu$ yield a simpler, but tighter, decision boundary.
For a dataset with a high number of outliers, too small of values of~$\nu$ can result 
in long training times and models that are overly sensitive to noise.

For a given OCSVM model, let~$n_{\out}$ be the number of outliers and~$n_{\inli}$ 
the number of inliers. We define
\begin{equation}\label{eq:OCSVM}
    r = \frac{n_{\out}}{n_{\inli}}
\end{equation}
If this ratio~$r$ changes over time, it is reasonable to assume that the underlying 
distribution of the data has changed. Therefore, we can use OCSVM to detect concept 
drift by monitoring the ratio of the number of outliers to the number of inliers.
Additional details on how we employ OCSVM for concept drift detection are 
given in Section~\ref{sect:detect}.
As far as the authors are aware, this is the first time that OCSVM has been used for 
concept drift detection, although OCSVMs have previously been used for 
malware anomaly detection~\cite{shi2024malwareocsvm}.

\subsubsection{Maximum Mean Discrepancy}\label{sec:mmd}

Concept drift can cause model performance degradation due to changes in the 
underlying data distribution. The authors of~\cite{rabanser2019datasetshift} provide
an empirical study of various distribution shift detection techniques, with the goal of better 
understanding the relationship between distribution shift and model performance degradation.

Given two sets of samples,  $x=\{x_1, x_2, \ldots, x_n\}$ 
and~$x'=\{x'_1, x'_2, \ldots, x'_m\}$, the goal
in distribution shift detection is to determine whether the 
probability distributions~$p(x)$ and~$q(x')$ are similar.
To detect whether two distributions are similar, we can use hypothesis testing 
with a two-sample test. Formally, we want to test the null hypothesis 
$$
    H_0: p(x) = q(x')
$$ 
against the alternative
$$
    H_A: p(x) \ne q(x')
$$ 
at a specified significance level~$\alpha$. We note in passing that distribution shift 
depends on the chosen representation of the samples, which we refer to as the
latent space.  
Depending on the latent space, 
two-sample tests can give different results.

Maximum Mean Discrepancy (MMD)~\cite{gretton2012kernel} is a popular kernel-based 
technique for multivariate two sample testing. MMD can distinguish between two 
distributions~$p$ and~$q$ based on the mean embeddings~$\mu_p$ and~$\mu_q$ 
in a special latent space (i.e., Hilbert space) that we denote as~$\hilbertSpace$.

MMD uses a kernel~$k$ to map samples onto~$\hilbertSpace$. 
The MMD is computed as
$$
    \mmd (\hilbertSpace, p, q) = \|\mu_p - \mu_q\|_{\hilbertSpace} .
$$
Given samples drawn from~$p$ and~$q$ we compute the unbiased estimator
$$
    \mmd^2 = \frac{1}{n^2} \sum_{i \neq j}^n k(x_i, x_j) 
    - \frac{2}{nm} \sum_{i=1}^{n} \sum_{j=1}^{m} k(x_i, x'_j)
    + \frac{1}{m^2} \sum_{i \neq j}^m k(x'_i, x'_j) 
$$
where~$k$ is the Gaussian exponential kernel, 
$k(x, \tilde{x}) = e^{{-\|x - \tilde{x}\|^2} / \sigma}$, 
and~$\sigma$ is the median distance between all pair of 
points~$(x, x')$~\cite{gretton2012kernel}.
The authors of~\cite{rabanser2019datasetshift} note that we 
can compute a~\pvalue\ using permutation tests on the resulting 
kernel matrix to perform hypothesis testing.
Once a~\pvalue\ is computed, we accept or reject the null hypothesis~$H_0$ 
at a chosen significance level~$\alpha$ depending on 
whether~$\mbox{\pvalue} > \alpha$ (accept~$H_0$)
or~$\mbox{\pvalue} \le \alpha$ (reject~$H_0$).
The Python package Alibi Detect~\cite{alibi-detect} contains implementations 
of all techniques described in~\cite{rabanser2019datasetshift}; we
use these implementations for our experiments.

\subsection{Learning Models}

In this section, we introduce the four learning models that are consider in our experiments.
These are standard learning models that serve to test our concept drift detection techniques.

\subsubsection{Multilayer Perceptron}

A MultiLayer Perceptron (MLP) is a classic type of feedforward Artificial Neural Network (ANN).
An MLP consists of an input layer, an output layer, and one or more hidden layers.
Each hidden layer is fully connected to the layers above and below.
MLPs are usually trained using backpropagation. 
Training MLPs is done in batches and epochs---one epoch represents a forward and 
backward pass through the entire training set, which is usually divided into minibatches.

The main hyperparameters of an MLP architecture are the number of hidden layers, 
the number of neurons in each layer, and the choice of activation functions. 
For classification tasks, cross-entropy is widely used, and it is available in 
\texttt{scikit-learn}~\cite{scikitLearnMLPClassifier}. 
For binary classification, Binary Cross Entropy (BCE), also known as Log Loss, 
is used~\cite{Terven2025}. Assuming a classifier is a probabilistic model, 
BCE measures the distance between two Bernoulli distributions---the true 
distribution and the predicted distribution given by the 
classifier~\cite{Terven2025}. 
Thus, for~$n$ samples, the~$\BCE$ is computed as
$$
    \BCE = -\frac{1}{n} \sum_{i=1}^{n} \big(y_i \log(p_i) + (1 - y_i) \log(1 - p_i)\big) 
$$
where~$y_i$ is the actual label (0 or~1) of sample~$i$, and~$p_i$ is the
predicted probability of the~$i^{\thth}$ sample being in class~1. 
Note that due to the~$\log$ term, BCE assigns a higher penalty to more confident, 
but incorrect, predictions.

\subsubsection{Support Vector Machine}

A Support Vector Machine (SVM) is a supervised machine learning model.
The goal of an SVM is to find the optimal hyperplane that separates the data into classes.

To achieve separability, SVMs can map the data into a higher dimensional space using 
a kernel function. The kernel function is selected such that computing 
the dot product in a higher dimensional space does not incur any significant 
computational cost in terms of the input data---this is known as the ``kernel trick.''
The most popular kernel functions are linear, polynomial, 
and radial basis function (rbf)~\cite{tabsharani2023svm}.
For additional details on SVMs, see, for example~\cite{cortes1995svm}.

\subsubsection{Random Forest}

Random Forest (RF) is an ensemble machine learning technique that combines 
decision trees to make a prediction. Random Forest uses ``bagging'' 
to create uncorrelated decision trees~\cite{ibm2021randomforest}.
Random Forest is a classical machine learning model that in practice can often rival 
more complex neural network models.

An important hyperparameter of RF is the depth of the trees.
A deeper depth means that more decisions are included in each tree, 
which can possibly lead to a higher accuracy, at the cost of a higher training time,
as well as an increased chance of overfitting. RF models tend to generalize well 
given the uncorrelated nature of the component decision trees.
For more details on RFs, refer to~\cite{breiman2001randomforests}.

\subsubsection{Extreme Gradient Boosting}

Similar to RF, eXtreme Gradient Boosting (XGB) is an ensemble machine learning 
technique based on decision trees. It is important to note that XGBoost is an optimized 
version of a general gradient boosting algorithm, with the goal of maximizing performance 
and efficiency~\cite{xgboostOriginal}.

XGBoost uses boosting instead of bagging, which distinguishes it from RF.
In practice, this means that while RF builds independent decision trees by 
random bagging of the features and samples, XGBoost builds decision trees sequentially.
In XGBoost, each new (possibly weak) classifiers aims to predict the errors of the 
previous classifier~\cite{lev2022xgboostArticle}. The intuition behind XGBoost is that each 
new classifier will generally be able to improve on the errors of the previous classifiers.

\section{Implementation}\label{chap:implementation}

In this section, we first discuss the dataset that we use in our experiments.
Then we present a detailed view of our experimental setup.

\subsection{Dataset}\label{sec:dataset}

The study of concept drift in malware detection requires a dataset in which samples 
can be placed on a relative timeline.  
We use the KronoDroid 
dataset, which was introduced in~\cite{guerra2021kronodroid}; we retrieved the 
dataset from~\cite{guerra2021github}.

The KronoDroid dataset contains~41,382 Android malware samples belonging 
to~240 distinct malware families. Note that the KronoDroid dataset contains both 
real and emulated samples---we only use real samples in this research.
Each sample in the dataset includes~200 static features 
(permissions, intents, hardware/software requirements, etc.) 
and~289 dynamic features (e.g., system calls).

According to the paper introducing KronoDroid~\cite{guerra2021kronodroid}, 
each sample contains four distinct timestamps: Earliest Modification, Last Modification, 
First Seen VT (on Virus Total) and First Seen ITW (in the wild).
However, in the dataset available from Github~\cite{guerra2021github}, 
malware samples contain only two timestamps: \texttt{EarliestModDate} and \texttt{HighestModDate}.
In this research, we use the \texttt{HighestModDate} timestamp as it is the best available 
timestamp in the dataset with the least number of invalid samples, 
according to~\cite{guerra2021kronodroid}.

\subsubsection{Malware Families}

Although there are~240 malware families in the KronoDroid dataset, we focus our attention 
on the five malware families with the largest number of samples.
By doing so, we can study concept drift and evaluate model performance 
for a significant number of samples over meaningful periods of time.
In this respect, our approach is analogous to that followed in~\cite{mishra2024cluster}.

The five malware families under consideration are the following.
\begin{description}
    \item[\texttt{Airpush/StopSMS}] is adware that displays unwanted ads and may silently 
    collect and forward user data~\cite{fsecure2025airpush}. Subsequently, we refer this family 
    simply as \texttt{Airpush}.
    \item[\texttt{SMSReg}] is riskware that, for example, may include a fake 
    ``Battery Improve'' application~\cite{fsecure2025smsreg}.
    \item[\texttt{Malap}] is spyware that collects sensitive information from a device~\cite{guerra2021kronodroid}.
    \item[\texttt{Boxer}] is a Trojan that pretends to be a legitimate installer but actually sends 
    premium-rate SMS messages, unbeknownst to the user~\cite{fsecure2025boxer}.
    \item[\texttt{Agent}] is a Trojan that downloads and installs adware or malware 
    onto a victim device~\cite{fsecure2025agent}.
\end{description}

\subsubsection{Data Preprocessing}

As mentioned above, each malware sample in our dataset includes~489 features,
200 of which are static and~289 of which are dynamic. However, some features are 
not numerical, but are instead text. As a preprocessing step, we remove all 
non-numerical features using the Polars library~\cite{polarsdatatypes}.
The dimension of each feature vector after this preprocessing is~470.

Additionally, we discard samples whose \texttt{HighestModDate} timestamp 
seems to be incorrect. In concrete terms, only samples with a timestamp in the 
format ``\texttt{MM/DD/YYYY}'' and years in the range of~2000 to~2025 are considered valid.
Table~\ref{tab:kronodroid_summary} summarizes the extracted samples from the KronoDroid dataset;
we explain the meaning of the ``batches'' column in Section~\ref{sect:defns}.

\begin{table}[!htb]
    \centering
    \caption{KronoDroid dataset after preprocessing} 
    \label{tab:kronodroid_summary}
    \begin{adjustbox}{scale=0.85}
        \begin{tabular}{l|r|c|c|c}
            \toprule
            \textbf{Family} & \textbf{Samples} & \textbf{Batches}  & \textbf{Earliest date} & \textbf{Latest date} \\
            \midrule
            \texttt{Agent}           & 2,826\zz\zz     &   \zz56     & 2008-02-28             & 2020-07-17           \\
            \texttt{Airpush} & 7,760\zz\zz        & 155     & 2008-02-29             & 2018-06-15           \\
            \texttt{Boxer}           & 3,597\zz\zz     & \zz71        & 2005-01-01             & 2018-06-15           \\
            \texttt{Malap}           & 4,018\zz\zz     & \zz80        & 2008-02-29             & 2020-11-11           \\
            \texttt{SMSreg}          & 4,989\zz\zz    & \zz99         & 2008-02-29             & 2020-11-09           \\
            \midrule
            Total           & 23,190\zz\zz      & 461     & 2005                    & 2020                  \\
            \bottomrule
        \end{tabular}
    \end{adjustbox}
\end{table}

\subsection{Experiment Setup}\label{sec:setup}

As discussed above, in this research we consider three scenarios to 
evaluate the benefits of employing concept drift 
detection in the malware domain. We refer to these three scenarios as static training, 
periodic retraining, and drift-aware retraining. In each scenario, we train each of the four 
machine learning classifiers discussed above
(MLP, SVM, RF, and XGB) on each of~20 distinct classification tasks.

Intuitively, we expect that periodic retraining will yield the highest accuracy while the static 
training accuracy will be the lowest. If we are able to accurately detect concept drift, 
then the drift-aware retraining accuracies should be comparable to the periodic 
retraining accuracies, while reducing the work required to retrain models.

\subsubsection{Definitions}\label{sect:defns}

Let ${\cal R} = \{\oc, \mk, \mmds\}$ denote the set of drift detection techniques considered. 
We let ${\cal L} = \{\mlp, \svm, \rf, \xgb\}$ be
the set of learning models under consideration.

Let~$F$ denote the samples belonging to a specific malware family.
For each experiment, let~$F_{d}$ be the samples of the malware family for 
which we want to detect where concept drift has occurred, and let~$F_{c}$ 
be the ``control'' samples, where~$F_{c}$ is a different malware family than~$F_{d}$.
For our binary classification experiments, $F_{d}$ and~$F_{c}$ are 
the classes that we train models to distinguish.
We employ the notation~${\cal F} = \{(F_{d},F_{c})\,|\,F_{d}\neq F_{c}\}$, 
and since we consider five families, $|{\cal F}|=20$.

For each malware family~$F$, we order the samples from oldest to most recent, 
based on the \texttt{HighestModDate} timestamp.
Let~$n$ be the number of samples in~$F$, and
denote the ordered samples of~$F$ as~$(f_{0}, f_{1}, \ldots, f_{n-1})$, 
where~$f_{i}$ precedes~$f_{j}$ 
in the temporal ordering whenever~$i < j$.

For a malware family~$F$, we partition the samples into consecutive temporal batches, 
truncating so that all temporal batches are of the same size.
Note that the smaller the batch size, the sooner we can detect concept drift,
but smaller batch sizes also tend to increase the variance, adding noise
to the process.
In all of our experiments, we define the batch size as~$b = 50$. Small-scale
tests showed that the results did not improve significantly for larger values of~$b$,
while adverse variance effects were observed for smaller batch sizes.

Assuming that~$|F|=n$, the number of batches in~$F$ is~$\lfloor n/50\rfloor$.
Letting~$B_i$ represent the~$i^{\thth}$ batch, we have
$$
    B_{i} = \{f_{i \cdot b}, f_{i \cdot b + 1}, \ldots, f_{(i+1) \cdot b - 1}\}
$$
Thus, $B_{0} = \{f_{0}, f_{1}, \ldots, f_{49}\}$, $B_{1} = \{f_{50}, f_{51}, \ldots, f_{99}\}$, 
and so on.

Let~$B_{i}^{\train}$ be the training subset and~$B_{i}^{\test}$ be the testing subset of~$B_i$.
We define~$B_{i}^{\train}$ to be the first~$t$ samples in~$B_{i}$ and $B_{i}^{\test}$ to be the 
remaining~$b - t$ samples in~$B_{i}$. We use~$t = 30$ samples for training and hence 
the number of samples used for testing per batch is~$b - t = 50 - 30 = 20$.
For example, for the first batch~$B_{0}$, we have~$B_{0}^{\train} = \{f_{0}, f_{1}, \ldots, f_{29}\}$ 
and~$B_{0}^{\test} = \{f_{30}, f_{31}, \ldots, f_{49}\}$.
The ``batches'' column in Table~\ref{tab:kronodroid_summary} provides 
the number of batches per family.

\subsubsection{Concept Drift Detection}\label{sect:detect}

As mentioned above, we consider two machine learning-based approaches for concept 
drift detection---one based on MK-Means, another based on OCSVM---as well as the statistical-based
MMD technique. In this section, we describe these three approaches in more detail.

\begin{description}
    \item[MK-Means]--- We employ the silhouette coefficient, based on MK-Means clustering, 
    to detect concept drift as follows. Using the notation above, we 
    let~$\matches{B}_0 = \{B_0, B_1\}$ and~$\matches{B}_1 = \{B_1, B_2\}$.
    Then we perform MK-Means clustering on the samples in~$\matches{B}_0$ and separately perform 
    MK-Means clustering on the samples in~$\matches{B}_1$. Next, we compute the average silhouette
    coefficient~$s_0$ for the clustering of~$\matches{B}_0$ and compute the average silhouette 
    coefficient~$s_1$ for the clustering of~$\matches{B}_1$.
    Then if~$|s_{1} - s_{0}| > \tau_{\mk}$, where~$\tau_{\mk}$ is a specified threshold, 
    we assume that concept drift has occurred at batch~$B_{1}$.
    We repeat this process for subsequent consecutive pairs of batches.
    \item[OCSVM]--- We employ OCSVM to detect concept drift as follows.
    We train an OCSVM on~$B_0$ and compute the ratio of 
    outliers to inliers---as per equation~\eref{eq:OCSVM}---which we denote as~$r_0$.
    The ratio~$r_0$ represents characteristics 
    of the data in~$B_0$. We then reuse this same OCSVM model to compute~$r_{1}$,
    the ratio of outliers to inliers over~$B_{1}$. 
    If~$|r_{1} - r_{0}| > \tau_{\oc}$, where~$\tau_{\oc}$ is a specified threshold, 
    then we assume that the samples in~$B_{1}$ differ from those in~$B_{0}$, 
    and hence concept drift has occurred. If concept drift is detected, we then retrain the 
    OCSVM on~$B_{1}$ and recompute $r_{1}$ on $B_{1}$ using this updated OCSVM
    model at the next step. On the other hand, if concept drift is not detected in~$B_{1}$, 
    we do not retrain the OCSVM. In either case, we then move on to consider~$B_{2}$, 
    and the process repeats.
    \item[MMD]--- We employ MMD to detect concept drift as follows.
    To identify whether concept drift has occurred at batch~$B_i$, 
    we let $B_{\reference} = B_{i-1}$ and $B_{\new} = B_{i}$.
    We then perform a two-sample test using MMD with a Gaussian kernel and 
    compute a~$p$-value for hypothesis testing as described in Section~\ref{sec:mmd}.
    Given a confidence level~$\alpha$, we let~$\tau_{\mmds} = \alpha$ to keep notation 
    homogenous across all drift detectors. If~$\mbox{$p$-value} \le \tau_{\mmds}$, we conclude 
    that drift has occurred at $B_i$; otherwise, no drift is detected.
\end{description}

\subsubsection{Training a model}

Consider a given pair of malware families which, as above, we denoted as~$(F_{d}, F_{c})$, 
with~$N$ batches~$B_i$ over~$F_{d}$. Let~$X^{\train}$ and~$X^{\test}$ be 
two disjoint subsets of~$F_{c}$. Let~$D_{i}^{\train} = (B_{i}^{\train} \cup X^{\train})$ 
and~$D_{i}^{\test} = (B_{i}^{\test} \cup X^{\test})$. This yields a labeled dataset, 
where the samples in~$B_i$ are one class, and the selected samples 
from~$F_{c}$, namely, $X^{\train}\cup X^{\test}$, are the other class.

A classification model~$M_i$ is trained on~$D_{i}^{\train}$ and evaluated 
on~$D_{i}^{\test}$\!\!. From Section~\ref{sect:defns}, 
we recall that~$|B_{i}^{\train}| = t$ and~$|B_{i}^{\test}| = b - t$.
Ideally, we want a balanced dataset with~$|X^{\train}| = t$ and~$|X^{\test}| = b - t$.
Therefore, in all experiments we let
$$
    X^{\train} = \{f^{c}_{0}, f^{c}_{1}, \ldots, f^{c}_{t-1}\}
$$
and
$$
    X^{\test} = \{f^{c}_{t}, f^{c}_{t+1}, \ldots, f^{c}_{b-1}\}
$$
As mentioned above, in all of our experiments, we let~$t = 30$ and~$b=50$, 
and hence~$b - t = 20$.

\subsubsection{Static Training}\label{sec:static_training}
    %
For a given drift family~$F_d$, 
control family~$F_c$, and learning model~$L\in{\cal L}$,
we train the model on~$D_0^{\train}$. We denote this trained model as~$M_0$. 
Then the accuracy~$A_{\static}(F_d,F_c,L)$ under the static training scenario is computed as
\begin{equation}\label{eq:static}
    A_{\static}(F_d,F_c,L) = \displaystyle\sum_{i=0}^{N-1} M_0(D_i^{\test}) \bigg/ 
    	\displaystyle\sum_{i=0}^{N-1} |D_i^{\test}|
\end{equation}
where~$M_0(D_i^{\test})$ is the number of correct predictions made by model~$M_0$ 
over~$D_i^{\test}$\!\!, and the drift family consists of~$N$ batches numbered~0 through~$N-1$.
This experiment represents the scenario where we train one model on the initial temporal batch, 
and use this model to classify the test samples in all subsequent batches.

\subsubsection{Periodic Retraining}\label{sec:periodic_retraining}
For a given drift family~$F_d$, control family~$F_c$, and learning model~$L$,
we train a model~$M_i$ on~$D_i^{\train}$ for each batch in the drift family~$F_{d}$.
The overall accuracy for this periodic retraining experiment is given by
\begin{equation}\label{eq:periodic}
    A_{\periodic}(F_d,F_c,L) = \displaystyle\sum_{i=0}^{N-1} M_i(D_i^{\test}) \bigg/ \displaystyle\sum_{i=0}^{N-1} |D_i^{\test}|
\end{equation}
where~$M_i(D_i^{\test})$ is the number of correct predictions made by~$M_i$ over~$D_i^{\test}$\!\!,
and the drift family consists of~$N$ batches numbered~0 through~$N-1$.

The periodic retraining experiment simulates the scenario where we train a new instance 
of the same model type on every batch of samples.
Note that it is not necessary that the hyperparameters of the model remain the 
same across batches, and hence we recompute the hyperparameters for each model~$M_i$.

\subsubsection{Drift-Aware Retraining}\label{sec:drift_aware_retraining}
The drift-aware retraining scenario is slightly more involved.
For a given drift family~$F_d$, control family~$F_c$, and learning model~$L$,
suppose that we detect concept drift at batches~$B_{i_1}$, $B_{i_2}$, $\ldots$, $B_{i_{\ell}}$, where
$$
    0 < i_1 < i_2 < \cdots < i_{\ell} < N-1
$$
with~$N$ being the number of batches in the drift family~$F_{d}$. 
Then defining~$i_0=-1$ 
and~$i_{\ell+1}=N-1$, 
we have
\begin{equation}\label{eq:drift}
    A_{\drift}(F_d,F_c,L,R) = \displaystyle\sum_{j=0}^{\ell}\,\sum_{k={i_j}+1\!\!\!\!}^{i_{j+1}}\! M_{{i_j}+1}(D_k^{\test})
    \bigg/ \displaystyle\sum_{i=0}^{N-1} |D_i^{\test}|
\end{equation}
%

\subsubsection{Efficiency}

The models trained in the drift-aware scenario are a subset of the 
models trained in the periodic retraining scenario.
This follows since the batches where concept drift is detected are a subset 
of all of the batches in~$F_{d}$, and in the periodic retraining scenario, 
we retrain a model at every batch of~$F_{d}$.
Consequently, one measure of efficiency for the drift-aware scenario is the percentage of 
batches where concept drift is detected---the more drift points detected, the more models 
that must be trained, resulting in a higher overall cost for training.
In this sense, the periodic retraining scenario has the worst possible efficiency,
and the static scenario has the best possible efficiency, since at least one model must be trained.

For specified families~$(F_d,F_c)$ and drift detection technique~$R$, let~$\ell$ be the
number of drift points detected. Then
we define the drift-aware training efficiency as 
\begin{equation}\label{eq:efficiency}
	{\cal E}(F_{d},F_{c},R) =
	\left\{
	\begin{array}{ll}
	1 - \displaystyle\frac{\ell}{N - 2} & \mbox{ if } R = \mk \\[1.5ex] 
	1 - \displaystyle\frac{\ell}{N - 1} & \mbox { otherwise}
	\end{array}
	\right.
\end{equation}
where~$N$ is the number of batches in~$F_{d}$.
Since our clustering approach uses pairs of batches, for the MK-Means 
drift detection technique we have~$N - 2$ possible drift points, whereas for
the other two drift detection techniques, we have~$N - 1$ possible drift points.

Note that by the efficiency measures in equation~\eref{eq:efficiency}, 
the static scenario will always achieve the maximum
efficiency of~1. However, in any case where concept drift occurs, we expect the static scenario to
have the lowest accuracy of our three experimental scenarios. Hence, the tradeoff between
efficiency and accuracy must be considered. We provide a detailed discussion of this 
inherent tradeoff in Section~\ref{sec:paretoFront}.

\subsubsection{Workflow}\label{sec:workflow}

The experiment workflow consists of preprocessing the dataset and then conducting the
experiments. The preprocessing step includes removing non-numerical features 
and discarding samples with obviously incorrect timestamps.
Details on the workflow for the experiments is given in Algorithm~\ref{alg:experiments}.

\begin{figure}[!htb]
    \renewcommand\theequation{{\color{blue}\arabic{equation}}}
    \algdef{SE}[SUBALG]{Indent}{EndIndent}{}{\algorithmicend\ }%
    \algtext*{Indent}
    \algtext*{EndIndent}
    \centering
    \begin{adjustbox}{scale=0.85}
        \begin{minipage}{0.9\textwidth}
            \begin{algorithm}[H]
                 \caption{Experiment workflow}\label{alg:experiments}
                \begin{algorithmic}[1]
                    \State \com{Step 0: Preliminaries}
                    \State \textbf{Input:} Select drift and control malware families $(F_{d}, F_{c})\in{\cal F}$
                    \State \textbf{\phantom{Input:}}  Let~$N = \mbox{number of batches in } F_{d}$
                    \State \textbf{\phantom{Input:}}  Select learning model $L\in{\cal L}=\{\mlp,\svm,\rf,\xgb\}$
                    \State \textbf{\phantom{Input:}}  Select drift detector $R\in{\cal R}=\{\oc,\mk,\mmds\}$
                    \State \textbf{\phantom{Input:}}  Specify drift detector threshold $\tau_R$
                    \State \textbf{Output:} Accuracy for static, periodic, and drift-aware scenarios
                    \State \textbf{\phantom{Output:}} Efficiency for drift-aware scenario 
                    \vspace{0.25cm}
                    \State \com{Step 1: Train models (including hyperparameter tuning)}
                    \For{$i=1,2,\ldots,N$}
                    \State Train model $M_i$ (of type $L$) on batch $B_i$ 
                    \EndFor
                    \vspace{0.15cm}
                    \State \com{Step 2: Static scenario}
                    \State Use trained $M_0$ to compute $A_{\static}(F_d,F_c,L)$ \com{equation~\eref{eq:static}}
                    \vspace{0.25cm}
                    \State \com{Step 3: Periodic retraining}
                    \State Use all models $M_i$ to compute $A_{\periodic}(F_d,F_c,L)$ \com{equation~\eref{eq:periodic}}
                    \vspace{0.25cm}
                    \State \com{Step 4: Drift-aware retraining}
                    \State Detect drift points $D$ using $R$ and threshold~$\tau_R$ \com{Section~\ref{sect:detect}}
                    \State Use~$D$ to compute accuracy $A_{\drift}(F_d,F_c,L,R)$ \com{equation~\eref{eq:drift}}
                    \State Compute efficiency ${\cal E}(F_d,F_c,R)$ \com{equation~\eref{eq:efficiency}}
                    \vspace{0.25cm}
                    \State \com{Step 5: Return results}
                    \State \textbf{Return:} $A_{\static}(F_d,F_c,L)$, $A_{\periodic}(F_d,F_c,L)$, 
                    			$A_{\drift}(F_d,F_c,L,R)$,
                    \State \phantom{\textbf{Return:}} and ${\cal E}(F_d,F_c,R)$
                \end{algorithmic}
            \end{algorithm}
        \end{minipage}
    \end{adjustbox}
\end{figure}

\subsection{Hyperparameters}

Before presenting our experimental results, we need to consider hyperparameters.
In this section, we first discuss hyperparameter tuning of the learning models, 
which affect the classification accuracy of each model.
After discussing hyperparameter tuning for our learning models,
we consider the hyperparameters for our drift detection techniques, 
which affect the tradeoff between accuracy and efficiency under the drift-aware scenario.

\subsubsection{Hyperparameter Tuning of Learning Models}

An important step in our classification experiments is to perform hyperparameter tuning 
to obtain optimized learning models.
We use Optuna~\cite{optunaHyperparameterDocs}, which performs 
a Tree-Structured Parzen Estimator (TPE)~\cite{watanabe2023tpe} search for the best 
hyperparameters. 

TPE is a Bayesian optimization algorithm.
Given an objective function~$f_{\theta}$ where~$\theta$ is a set of parameters, 
TPE performs multiple trials over~$\theta$, iteratively searching for optimal values.
This approach is different from a grid search because TPE does not try every 
value defined in the space---it uses Bayesian estimates to find better regions.
We use a total of~100 trials per model.

The set of hyperparameters considered for each ML model
are given in Table~\ref{tab:model_hyperparameter_tuning}.
Here, we employ the notation~$(a, b)$ to indicate that the open interval 
from~$a$ to~$b$ was specified for Optuna, while
the notation~$\{x, y\}$ denotes a discrete set of values.

\begin{table}[!htb]
    \centering
    \vspace{0.5cm}
    \caption{Hyperparameters for learning models}\label{tab:model_hyperparameter_tuning}
    \begin{adjustbox}{scale=0.85}
        \begin{tabular}{l|c|c}
            \toprule
            \textbf{Model}       & \textbf{Hyperparameters} & \textbf{Search space}         \\
            \midrule
            \multirow{2}{*}{SVM} & $c$ (regularization)     & $(e^{-4}, e^{4})$             \\
                                 & $\kernel$                & \{rbf\}                       \\
            \midrule
            \multirow{2}{*}{RF}  & $\nestimators$           & $(10,500)$                     \\
                                 & $\mxD$                   & $(5, 100)$                    \\
            \midrule
            \multirow{3}{*}{XGB} & $\nestimators$           & $(10, 300)$                     \\
                                 & $\mxD$                   & $(5, 15)$                    \\
                                 & $\learningrate$          & $\{0.01, 0.05, 0.1, 0.2, 0.3\}$          \\
            \midrule
            \multirow{3}{*}{MLP} & $\nunits$            & $(8, 256)$                    \\
                                 & $\activation$            & $\{\relu, \tanh, \logistic\}$ \\
                                 & $\solver$                & $\{\adam, \sgd\}$             \\
            \bottomrule
        \end{tabular}
    \end{adjustbox}
\end{table}

Given~$(F_{d}, F_{c}, L)$, where~$L$ is the selected type of learning model,
the objective function of each model~$M_i$ is defined as
\begin{equation}\label{eq:miObjectiveFunction}
    f^{\theta}_i = \frac{M_i^{\theta}(D_i^{\test})}{|D_i^{\test}|}
\end{equation}
where~$M_i^{\theta}$ is initialized using parameters~$\theta$, with the model 
trained on~$D_i^{\train}$, and~$M^{\theta}_i(D_i^{\test})$ denotes the number of correct 
predictions of~$M^{\theta}_i$ on~$D_i^{\test}$,
that is, we are optimizing each model based on the test accuracy.
All other parameters needed for a model~$M_i$ not mentioned in the table 
are set to their default values in the respective libraries.
Since we tune the hyperparameters of 
one model~$M_i$ per~$(F_{d},F_{c}, L)$ combination and per temporal batch,
the selected hyperparameters need not be the same.
For all models, we use early stopping---if a set of hyperparameters 
gives an objective function with ideal separation, we stop searching the space.

For each drift family, there are four~$(F_{d}, F_{c})$ pairs,
we have~461 total temporal batches (accounting for all drift families), 
and there are four machine learning models.
Therefore, we have~$4 \cdot 4 \cdot 461 = 7{,}376$ combinations.
Furthermore, for each of these combinations, we run~100 trials,
which gives us a total number of~$(F_{d}, F_{c}, L)$ combinations 
for which hyperparameters are determined of
$$
  100 \cdot 7{,}376 = 737{,}600 .
$$

We note in passing that the standard approach would be to tune the hyperparameters using a separate validation set. 
However, because of the small number of samples, such a validation set is not available. Tuning on the training 
set would also not be ideal because the reported metrics are computed on the test set, and overfitting is likely, 
especially given the limited sample size.
Thus, the use of test-set tuning is intentional and can be interpreted as an oracle-style upper bound 
on performance rather than as a realistic deployment procedure. The goal of our experimental design is to 
initially establish the two baseline scenarios (static and periodic retraining), and then compare these two
to various concept drift detection strategies under the drift-aware scenario. Under all three of these scenarios, 
the hyperparameter tuning is conducted in a similar fashion, and hence any change in accuracy 
in the drift-aware scenario should be attributable to the drift detection mechanism.

\subsubsection{Hyperparameters for Drift Detectors}\label{sec:drift_hyperparameter_tuning}

It is clear from our experimental design that if all batches are detected as drift points, 
then the drift-aware scenario is identical to the periodic retraining scenario.
Furthermore, in equation~\eref{eq:efficiency}
we define efficiency so that the larger the number of detected drift points, 
the lower the efficiency in the drift-aware scenario.
In practice, we would want to balance accuracy and efficiency.
Since the hyperparameters of the drift detectors determine their respective 
detection processes, the tradeoff between accuracy and efficiency can be analyzed 
as a function of these hyperparameters.

We can model this problem as a multi-objective optimization 
of conflicting objectives. This means that an increase in one objective 
produces a decrease in the other objective.
Given a drift detector~$R\in {\cal R}$, we denote its hyperparameter space as~$\Omega_R$.
The drift-aware accuracy~$A_{\drift}$ and retraining efficiency~${\cal E}$ are the objective functions.
Note that hyperparameter tuning of the models~$M_i$ is orthogonal to the 
hyperparameter tuning of the drift detector, and hence we do not need to retrain the~$M_i$ models.
Therefore, we can simply rerun Step~4 (lines 16--19) of Algorithm~\ref{alg:experiments}
using a grid search over~$\Omega_R$ to obtain multiple values for~$A_{\drift}$ and~${\cal E}$.
Table~\ref{tab:drift_detection_hyperparameters} shows the hyperparameter space 
that we use in a grid search for our OCSVM, MK-Means, and MMD concept drift detection 
techniques. We test all learning models and family combinations, and
consequently, we run Step~4 of Algorithm~\ref{alg:experiments} a total of
$$
    4 \cdot 20 \cdot (540 + 240 + 100) = 70{,}400
$$
times.

\begin{table}[!htb]
    \centering
    \caption{Hyperparameters for each drift detector}\label{tab:drift_detection_hyperparameters}
    \begin{adjustbox}{scale=0.85}
        \begin{tabular}{l|c|c|c}
            \toprule
            \multirow{2}{*}{\textbf{Drift detector~$R$}}  & \multicolumn{3}{c}{$\Omega_R$} \\ \cline{2-4} 
            & \textbf{Hyperparameters\vphantom{$N^{N^N}$}} & \textbf{Values tested}   & \textbf{Total values}     \\
            \midrule
            \multirow{2}{*}{MK-Means} & clusters                 & $\{2, 4, 6, 8\}$  
            			 & \multirow{2}{*}{$4 \cdot 60 = 240$} \\
                                      & $\tau_{\mk}$                   & $\{0.01, 0.02, \ldots, 0.6\}$ \\
            \midrule
            \multirow{2}{*}{OCSVM}    & $\nu$                    & $\{0.1, 0.2, \ldots, 0.9\}$  
            			& \multirow{2}{*}{$9 \cdot 60 = 540$} \\
                                      & $\tau_{\oc}$                   & $\{0.01, 0.02, \ldots, 0.6\}$ \\
            \midrule
            MMD                       & $\tau_{\mmds} = \alpha$          & $\{0.001, 0.002,\ldots, 0.1\}$  & 100\\
            \bottomrule
        \end{tabular}
    \end{adjustbox}
\end{table}

\subsection{Pareto Front}\label{sec:paretoFront}

According to Algorithm~\ref{alg:experiments}, in the drift-aware
scenario, the accuracy~$A_{\drift}$ depends on~$F_{d}$, $F_{c}$, $L$, and~$R$,
while the efficiency~${\cal E}$ depends only on~$F_{d}$, $F_{c}$, and~$R$. However, this is
not entirely correct, as both accuracy and efficiency depend on the 
selected hyperparameters~$\omega\in\Omega_R$ where, as above, $\Omega_R$ 
is a set of hyperparameters corresponding to drift detector~$R$.
Furthermore, there is an inherent tradeoff between accuracy and
efficiency. One technique for selecting the hyperparameters~$\omega$
is to find Pareto efficient solutions, also known as a 
Pareto Front~\cite{dichotomousPreferences2005}.

Pareto Fronts are widely used in engineering and economics where 
tradeoffs have to be made between two conflicting objectives~\cite{Coello2007},
such as the tradeoff between fuel efficiency and engine power.
By using a Pareto Front, it is possible to find a set of points where 
optimal tradeoffs can be made---in a sense that is made clear below.
The Pareto Front is also referred to as the set of non-dominated solutions.

Formally, we consider a compact parameter space~$\mathbb{R}^n$ and an 
objective space~$\mathbb{R}^m$ with a mapping function~$f$ such that
$$
    f : X \to \mathbb{R}^m
$$
where~$X \subseteq \mathbb{R}^n$.
Then, the objective values can be represented as a set of vectors
$$
    Y = \{\, y \in \mathbb{R}^m \mid y = f(x),\ x \in X \} .
$$
Let $\succ$ be a preference operator that dictates whether we want to maximize or 
minimize the objective. A vector~$y'' \in \mathbb{R}^m$ is said to strictly dominate 
another vector~$y' \in \mathbb{R}^m$ if
$$
    y'' \succ y'.
$$
Then the Pareto Front can be expressed as
$$
    P(Y) = \{\, y' \in Y \mid \{\, y'' \in Y \mid y'' \succ y',\ y'' \neq y' \,\} = \varnothing \} .
$$
Intuitively, the points belonging to the Pareto Front create a boundary that surrounds 
all other points of the objective. Points forming the boundary have the best possible 
tradeoff, in the sense that moving inside the boundary produces a negative effect 
on at least one of the objectives.

For a drift detector~$R$ and its parameter space~$\Omega_R$, 
we perform a Pareto Front analysis per~$(F_{d}, F_{c}, L)$ combination,
that is, for each combination of drift family, control family, and learning model.
Suppressing the dependence on~$(F_{d}, F_{c}, L)$,
as the objectives, we use~$(\Delta_{\ddrift}, {\cal E})$, where
$$
    \Delta_{\ddrift} = A_{\drift} - A_{\static} .
$$
Note that we use~$\Delta_{\ddrift}$ instead of~$A_{\drift}$ in the objective because it better conveys 
the desired optimization, which is to maximize the difference from the static baseline.
Of course, given~$\Delta_{\ddrift}$ and the static baseline~$A_{\static}$, 
we can trivially determine~$A_{\drift}$.

We use the \texttt{Paretoset} Python package, which implements the Skyline Query operator~\cite{maximaVectors1975,skylineQueries2015} to determine the Pareto Front for 
a set of points in the objective space. We note in passing that finding the Pareto Front 
without a set of points in the objective space can be achieved through evolutionary methods~\cite{Coello2007,evolutionaryMultiobjective2007} which fall under multi-objective 
optimization algorithms. We do not require a multi-objective optimization algorithm
since a grid search generates the set of points from which
we can directly determine the Pareto Front.

\subsection{Consolidating Drift-Aware Experiments}\label{sec:consolidatingDriftExperiments}

We now discuss how to use Pareto Front analysis to obtain meaningful results for 
comparing our drift detection techniques to the static and periodic scenarios. 
Again, this additional analysis is required
since it is necessary to tune various hyperparameters of each drift detection
technique, and when doing so, there is an inherent tradeoff between accuracy and efficiency.


Recall that~$L\in{\cal L}$ is the learning model, $R\in{\cal R}$ is the drift detection technique, 
$\Omega_R$ set of all hyperparameters tested for~$R$, while~$\omega$ 
represents a specific selection of hyperparameters from~$\Omega_R$.
Also, $F_{d}$ is the drift family, and~$F_{c}$ is the control family, 
with~$(F_{d},F_{c})\in{\cal F}$.
We require the additional notation
\begin{align*}
  \Omega^{\fame}_R &= \mbox{set of hyperparameters tested corresponding to Pareto Front} \\
  \omega^{\fame} &= \mbox{specific selection of hyperparameters from~$\Omega^{\fame}_R$} 
\end{align*}
Since we have five malware families, 
$|{\cal F}| = 20$, and
the values of~$|\Omega_R|$ 
are given in Table~\ref{tab:drift_detection_hyperparameters}.

\subsubsection{Mean Static and Periodic Accuracies}

In the static scenario, the accuracy depends only on~$F_{d}$, $F_{c}$, and~$L$,
and hence we denote each of these accuracy values as~$A_{\static}(F_{d},F_{c},L)$.  
Then the average accuracy per learning model~$L$ is given by
$$
    {\cal A}_{\static}(L) = \frac{1}{|{\cal F}|} \sum_{(F_{d},F_{c})\in{\cal F}} A_{\static}(F_{d},F_{c},L)
$$
Similarly, in the periodic scenario, the average accuracy
per learning model is
$$
    {\cal A}_{\periodic}(L) = \frac{1}{|{\cal F}|} \sum_{(F_{d},F_{c})\in{\cal F}} 
    	A_{\periodic}(F_{d},F_{c},L)
$$

\subsubsection{Mean Drift-Aware Accuracy}\label{sect:Bda}

The drift-aware scenario is more complex. As in the static and periodic
scenarios, the accuracy depends 
on~$F_{d}$, $F_{c}$, and~$L$. However,
in the drift-aware scenario, the accuracy also depends on the drift detection
technique~$R$ and, more specifically,
the selected hyperparameters~$\omega\in\Omega_R$.
Hence, in this scenario,
we denote the accuracy as~$A_{\drift}(F_{d},F_{c},L,R,\omega)$.

For a given~$\omega\in\Omega_R$, we can average over all~$(F_{d},F_{c})\in{\cal F}$ to obtain
\begin{equation}\label{eq:dF}
  A_{\drift}^{{\cal F}}(L,R,\omega) 
  	= \frac{1}{|{\cal F}|}\sum_{(F_{d},F_{c})\in{\cal F}} A_{\drift}(F_{d},F_{c},L,R,\omega)
\end{equation}
On the other hand, for a given~$(F_{d},F_{c})\in{\cal F}$, 
we can average over all~$\omega\in\Omega_R$ to obtain
\begin{equation}\label{eq:dO}
  A_{\drift}^{\Omega_R}(F_d,F_c,L,R) 
  	= \frac{1}{|\Omega_R|}\sum_{\omega\in\Omega_R} A_{\drift}(F_{d},F_{c},L,R,\omega)
\end{equation}
From equation~\eref{eq:dF}, the mean accuracy per~$(L,R)$ pair can be computed as
\begin{equation}\label{eq:AdaLR1}
  {\cal A}^{x}_{\drift}(L,R) = \frac{1}{|\Omega_R|}\sum_{\omega\in\Omega_R} 
  A_{\drift}^{{\cal F}}(L,R,\omega)
\end{equation}
Alternatively, based on equation~\eref{eq:dO}, we can compute the mean accuracy per~$(L,R)$ pair as
\begin{equation}\label{eq:AdaLR2}
  {\cal A}^{y}_{\drift}(L,R) = \frac{1}{|{\cal F}|}\sum_{(F_{d},F_{c})\in{\cal F}}
  A_{\drift}^{\Omega_R}(F_d,F_c,L,R)
\end{equation}
It is clear that
$$
  {\cal A}^{x}_{\drift}(L,R) = {\cal A}^{y}_{\drift}(L,R)
$$
and we denote this common value as~${\cal A}_{\drift}(L,R)$.
The calculations of~${\cal A}_{\drift}(L,R)$, using either equation~\eref{eq:AdaLR1}
or~\eref{eq:AdaLR2}, is illustrated in Table~\ref{fig:comp_A}.

\begin{table}[!htb]
\centering
\caption{Compute~${\cal A}_{\drift}$ as average of final row or final column ($N=|{\cal F}|$)}
	\label{fig:comp_A}
\scalebox{0.685}{$
\def\arraystretch{1.175}
\begin{array}{ccccc|c}\toprule
A(F_{d_0},F_{c_0},\omega_0) 
	& A(F_{d_0},F_{c_0},\omega_1) 
	& A(F_{d_0},F_{c_0},\omega_2) 
	& \cdots
	& A(F_{d_0},F_{c_0},\omega_{|\Omega|-1}) 
	& A_{\drift}^{\Omega}(F_{d_0},F_{c_0}) \\ 
A(F_{d_1},F_{c_1},\omega_0) 
	& A(F_{d_1},F_{c_1},\omega_1) 
	& A(F_{d_1},F_{c_1},\omega_2) 
	& \cdots
	& A(F_{d_1},F_{c_1},\omega_{|\Omega|-1}) 
	& A_{\drift}^{\Omega}(F_{d_1},F_{c_1}) \\
A(F_{d_2},F_{c_2},\omega_0) 
	& A(F_{d_2},F_{c_2},\omega_1) 
	& A(F_{d_2},F_{c_2},\omega_2) 
	& \cdots
	& A(F_{d_2},F_{c_2},\omega_{|\Omega|-1}) 
	& A_{\drift}^{\Omega}(F_{d_2},F_{c_2}) \\
\vdots
	& \vdots 
	& \vdots 
	& \vdots 
	& \vdots 
	& \vdots \\
A(F_{d_{N-1}},F_{c_{N-1}},\omega_0) 
	& A(F_{d_{N-1}},F_{c_{N-1}},\omega_1) 
	& A(F_{d_{N-1}},F_{c_{N-1}},\omega_2) 
	& \cdots
	& A(F_{d_{N-1}},F_{c_{N-1}},\omega_{|\Omega|-1}) 
	& A_{\drift}^{\Omega}(F_{d_{N-1}},F_{c_{N-1}}) \\ \midrule
A_{\drift}^{\cal F}(\omega_0) 
	& A_{\drift}^{\cal F}(\omega_1) 
	& A_{\drift}^{\cal F}(\omega_2) 
	& \cdots 
	& A_{\drift}^{\cal F}(\omega_{|\Omega|-1}) 
	& {\cal A}_{\drift}\\
\bottomrule
\end{array}
$}
\end{table}

\subsubsection{Mean Drift-Aware Accuracy over Pareto Front}\label{sect:BdaPF}

Computing of~${\cal A}_{\drift}(L,R)$ using either equation~\eref{eq:AdaLR1} 
or~\eref{eq:AdaLR2}
treats all hyperpameter combinations~$\omega\in\Omega_R$
the same. This is clearly not ideal, as many hyperparameter values will yield suboptimal 
results. To mitigate this issue, we now consider
an approach analogous to that in Section~\ref{sect:Bda},
but restricted to points on the relevant Pareto Front.

For each~$\omega\in\Omega_R$, we first 
determine~$A_{\drift}^{{\cal F}}(L,R,\omega)$ 
using equation~\eref{eq:dF}. Then we compute
\begin{equation}\label{eq:X}
  {\cal A}^{x\kern-1pt\fame}_{\drift}(L,R) = \frac{1}{|\Omega^{\fame}_R|}\sum_{\omega^{\fame}\in\Omega^{\fame}_R}
  A_{\drift}^{{\cal F}}(L,R,\omega^{\fame})
\end{equation}
where~$\Omega^{\fame}_R$ is the set of hyperparameter values 
on the Pareto Front that correspond 
to~$A_{\drift}^{{\cal F}}(L,R,\omega)$. 
Note that this computation 
of~${\cal A}^{x\kern-1pt\fame}_{\drift}(L,R)$
is analogous to that of~${\cal A}^{x}_{\drift}(L,R)$ in equation~\eref{eq:AdaLR1},
but restricted to the Pareto Front of~$A_{\drift}^{{\cal F}}(L,R,\omega)$.
The calculation of~${\cal A}^{x\kern-1pt\fame}_{\drift}(L,R)$ is illustrated in Table~\ref{fig:comp_X}.

\begin{table}[!htb]
\centering
\caption{Compute~${\cal A}^{x\kern-1pt\fame}_{\drift}$ as average of~$A_{\drift}^{\cal F}(\omega^{\fame}_i)$}
\label{fig:comp_X}
\scalebox{0.75}{$
\def\arraystretch{1.175}
\begin{array}{ccccc|c}\toprule
A(F_{d_0},F_{c_0},\omega_0) 
	& A(F_{d_0},F_{c_0},\omega_1) 
	& A(F_{d_0},F_{c_0},\omega_2) 
	& \cdots
	& A(F_{d_0},F_{c_0},\omega_{|\Omega|-1}) \\
A(F_{d_1},F_{c_1},\omega_0) 
	& A(F_{d_1},F_{c_1},\omega_1) 
	& A(F_{d_1},F_{c_1},\omega_2) 
	& \cdots
	& A(F_{d_1},F_{c_1},\omega_{|\Omega|-1}) \\
A(F_{d_2},F_{c_2},\omega_0) 
	& A(F_{d_2},F_{c_2},\omega_1) 
	& A(F_{d_2},F_{c_2},\omega_2) 
	& \cdots
	& A(F_{d_2},F_{c_2},\omega_{|\Omega|-1}) \\
\vdots
	& \vdots 
	& \vdots 
	& \vdots 
	& \vdots \\
A(F_{d_{N-1}},F_{c_{N-1}},\omega_0) 
	& A(F_{d_{N-1}},F_{c_{N-1}},\omega_1) 
	& A(F_{d_{N-1}},F_{c_{N-1}},\omega_2) 
	& \cdots
	& A(F_{d_{N-1}},F_{c_{N-1}},\omega_{|\Omega|-1}) \\ \midrule
A_{\drift}^{\cal F}(\omega_0) 
	& A_{\drift}^{\cal F}(\omega_1) 
	& A_{\drift}^{\cal F}(\omega_2) 
	& \cdots 
	& A_{\drift}^{\cal F}(\omega_{|\Omega|-1}) \\
\multicolumn{5}{c|}{\raisebox{.5\normalbaselineskip}[0pt][0pt]{%
	$\underbrace{\hspace*{0.875\textwidth}}$}} & {\cal A}^{x\kern-1pt\fame}_{\drift} \\
\multicolumn{5}{c|}{\mbox{Determine Pareto Front }\Omega^{\fame}\mbox{ from } 
	\{A_{\drift}^{\cal F}(\omega_i)\}_{i=0}^{|\Omega|-1}} \\
\bottomrule
\end{array}
$}
\end{table}

Alternatively, we can first sum over the Pareto Front points 
for each combination of~$(F_{d},F_{c},L,R)$. 
Analogous to equation~\eref{eq:dO}, in this case we compute
$$
  A_{\drift}^{\Omega^{\fame}_R}(F_d,F_c,L,R) 
  	= \frac{1}{|\Omega^{\fame}_R|}
		\sum_{\omega^{\fame}\in\Omega^{\fame}_R} A_{\drift}(F_{d},F_{c},L,R,\omega^{\fame})
$$
where~$\Omega^{\fame}_R$ is the set of hyperameter values that
correspond to Pareto Front points of~$(F_{d},F_{c},L,R)$.
Then averaging~$A_{\drift}^{\Omega^{\fame}_R}(F_d,F_c,L,R)$ 
over all~$(F_{d},F_{c})\in{\cal F}$, we obtain
\begin{equation}\label{eq:Y}
    {\cal A}^{y\kern-1pt\fame}_{\drift}(L,R) 
    	= \frac{1}{|{\cal F}|} \sum_{(F_{d},F_{c})\in{\cal F}} 
	    A_{\drift}^{\Omega^{\fame}_R}(F_d,F_c,L,R)
\end{equation}
Note that this computation of~${\cal A}^{y\kern-1pt\fame}_{\drift}(L,R)$
is analogous to that in equation~\eref{eq:AdaLR2},
but restricted to the relevant Pareto Front values.
The calculation of~${\cal A}^{y\kern-1pt\fame}_{\drift}(L,R)$ is illustrated in Table~\ref{fig:comp_Y}.

\begin{table}[!htb]
\centering
\caption{Compute~${\cal A}^{y\kern-1pt\fame}_{\drift}$ as average of final column ($N=|{\cal F}|$)}\label{fig:comp_Y}
\scalebox{0.675}{$
\def\arraystretch{1.175}
\begin{array}{ccccc|c}\toprule
A(F_{d_0},F_{c_0},\omega^{\fame}_0) 
	& A(F_{d_0},F_{c_0},\omega^{\fame}_1) 
	& A(F_{d_0},F_{c_0},\omega^{\fame}_2) 
	& \cdots
	& A(F_{d_0},F_{c_0},\omega^{\fame}_{|\Omega^{\fame}|-1}) 
	& A_{\drift}^{\Omega^{\fame}}(F_{d_0},F_{c_0}) \\ 
A(F_{d_1},F_{c_1},\omega^{\fame}_0) 
	& A(F_{d_1},F_{c_1},\omega^{\fame}_1) 
	& A(F_{d_1},F_{c_1},\omega^{\fame}_2) 
	& \cdots
	& A(F_{d_1},F_{c_1},\omega^{\fame}_{|\Omega^{\fame}|-1}) 
	& A_{\drift}^{\Omega^{\fame}}(F_{d_1},F_{c_1}) \\
A(F_{d_2},F_{c_2},\omega^{\fame}_0) 
	& A(F_{d_2},F_{c_2},\omega^{\fame}_1) 
	& A(F_{d_2},F_{c_2},\omega^{\fame}_2) 
	& \cdots
	& A(F_{d_2},F_{c_2},\omega^{\fame}_{|\Omega^{\fame}|-1}) 
	& A_{\drift}^{\Omega^{\fame}}(F_{d_2},F_{c_2}) \\
\vdots
	& \vdots 
	& \vdots 
	& \vdots 
	& \vdots 
	& \vdots \\
A(F_{d_{N-1}},F_{c_{N-1}},\omega^{\fame}_0) 
	& A(F_{d_{N-1}},F_{c_{N-1}},\omega^{\fame}_1) 
	& A(F_{d_{N-1}},F_{c_{N-1}},\omega^{\fame}_2) 
	& \cdots
	& A(F_{d_{N-1}},F_{c_{N-1}},\omega^{\fame}_{|\Omega^{\fame}|-1}) 
	& A_{\drift}^{\Omega^{\fame}}(F_{d_{N-1}},F_{c_{N-1}}) \\ \midrule
	& 
	&  
	&  
	& 
	& {\cal A}^{y\kern-1pt\fame}_{\drift}\\
\bottomrule
\end{array}
$}
\end{table}

When considering all of the hyperparameter combinations~$\omega\in\Omega_R$,
equations~\eref{eq:AdaLR1} and~\eref{eq:AdaLR2} 
show that the order of summation can be interchanged. 
However, when restricting our attention to hyperparameter values of~$R$
corresponding to Pareto Front points, this is no longer the case---in general,
there is no reason to expect that~${\cal A}^{x\kern-1pt\fame}_{\drift}(L,R)$ will yield the
same result as~${\cal A}^{y\kern-1pt\fame}_{\drift}(L,R)$.

\subsubsection{Representative Accuracy over Pareto Front}

In practice, we would need to select a specific value for the hyperparameters of the drift
detection technique~$R$, and we would almost certainly want to select such
a value from the Pareto Front. 
In this section we define such a representative value that is, in a sense, near the ``middle''
and, therefore, we refer to it as the median.

For a give pair of malware families~$(F_d,F_c)$, learning model~$L$, and drift detection
technique~$R$,
let~$(x_{\min},y_{\min})$ be the point on the Pareto Front that gives the minimum $\Delta\mbox{Accuracy}$
and let~$(x_{\max},y_{\max})$ be the point on the Pareto Front where the maximum $\Delta\mbox{Accuracy}$ is attained.
Then let
$$
  x_a = \frac{x_{\min}+x_{\max}}{2} \mbox{\ \ and\ \ } y_a = \frac{y_{\min}+y_{\max}}{2}
$$
The equation of the line from the origin through~$(x_a,y_a)$ is given by
\begin{equation}\label{eq:median}
  y = \frac{y_a}{x_a}\, x
\end{equation}

Let~$\omega^{\fame}_m\in\Omega^{\fame}_R$ to be the
hyperparameter values corresponding to the point on the Pareto Front that is
closest to the line in equation~\eref{eq:median}. Then we define the median accuracy 
for a given~$(L,R)$ pair as
$$
  {\cal A}^{m\kern-1pt\fame}_{\drift}(L,R) = \frac{1}{|{\cal F}|}\sum_{(F_d,F_c)\in {\cal F}} 
  	A_{\drift}(F_{d},F_{c},L,R,\omega^{\fame}_m)
$$

For each~$(L,R)$ pair, we now have four distinct measures of the 
accuracy, namely,
\begin{equation}\label{eq:fourA}
  {\cal A}_{\drift}(L,R),\  {\cal A}^{x\kern-1pt\fame}_{\drift}(L,R),\  {\cal A}^{y\kern-1pt\fame}_{\drift}(L,R),
  	\mbox{\ and\ } {\cal A}^{m\kern-1pt\fame}_{\drift}(L,R)
\end{equation}
Of course, for each of the accuracy measures in equation~\eref{eq:fourA} there is a 
corresponding efficiency that we denote as
\begin{equation}\label{eq:fourE}
  {\cal E}_{\drift}(L,R),\  {\cal E}^{x\kern-1pt\fame}_{\drift}(L,R),\  {\cal E}^{y\kern-1pt\fame}_{\drift}(L,R),
  	\mbox{\ and\ } {\cal E}^{m\kern-1pt\fame}_{\drift}(L,R)
\end{equation}
respectively.

\subsection{Software and Development Environment}\label{sec:environment}

Table~\ref{tab:libraries_purpose} summarizes the libraries and software used in this research.
Note that we primarily use Python for the experiments. In addition,
we use Docker to run a \texttt{PostgreSQL} container and a volume to store the hyperparameter 
tuning trial information generated by Optuna~\cite{optunaHyperparameterDocs}.
The best models $M_i$ are stored on disk as~\texttt{.pkl} files.
We export hyperparameter tuning information and experiment results as~\texttt{.csv} 
files for later analysis using \texttt{Matplotlib}. 
All experiments were run on a CPU machine; no GPUs were employed.
To allow for reproducibility, any machine learning model, function, or library used in this research 
that requires a random seed as input has been set to a fixed value of~420.

\begin{table}[!htb]
    \centering
    \caption{Software summary}
    \label{tab:libraries_purpose}
    \begin{adjustbox}{scale=0.85}
        \begin{tabular}{c|l}
            \toprule
            \textbf{Software}     & \hspace*{0.75in}\textbf{Purpose}                                             \\ \midrule
            Python3.10                    & Main programming language used 
            \\ \midrule
            Poetry                        & Python dependency management and packaging tool
            \\ \midrule
            \multirow{2}{*}{Polars}       & Loading and preprocess KronoDroid dataset             \\
                                          & Type inference and data manipulation
            \\ \midrule
            Matplotlib                    & Build and visualize graphs and plots
            \\ \midrule
            NumPy                         & Numerical computations and array manipulations
            \\ \midrule
            Alibi-Detect                  & Implementation for MMD
            \\ \midrule
            Pytorch                       & Backend needed to run MMD
            \\ \midrule
            Paretoset                     & Find the Pareto Front
            \\ \midrule
            Optuna                        & Automatic hyperparameter optimization framework
            \\ \midrule
            \multirow{2}{*}{\texttt{scikit-learn}} & OCSVM and MK-Means implementations                           \\
                                          & Implementations of MLP, SVM, and RF classifiers
            \\ \midrule
            XGBoost                       & XGB implementation
            \\ \midrule
            Docker                        & Postgres container for storing hyperparameters
            \\ \midrule
            Psycopg2                      & PostgreSQL database adapter for Python
            \\ \bottomrule
        \end{tabular}
    \end{adjustbox}
\end{table}

\section{Experimental Results and Analysis}\label{chap:results}

In this section, we present the results for the concept drift detection experiments discussed above. 
The primary goal of these experiments is to determine the accuracy and efficiency of each drift detector
(MK-Means, OCSVM, and MMD), as compared to the static and periodic retraining scenarios. 

\subsection{Static and Periodic Results}\label{sec:staticPeriodicResults}

Recall that our static and periodic experiments are independent of any drift detection 
technique. In subsequent sections, we consider accuracy and efficiency in the
drift-aware scenario based on each of the three drift detectors under consideration.

Figures~\ref{fig:periodicStatic_MLP} through~\ref{fig:periodicStatic_XGB}
in the Appendix
provides bar graphs of the static accuracies~$A_{\static}$ 
and periodic accuracies~$A_{\periodic}$.
These results are provided for all four learning models~$L$ 
and all~20 family combinations for~$(F_{d}, F_{c})$.

From Figures~\ref{fig:periodicStatic_MLP} through~\ref{fig:periodicStatic_XGB}
we observe that MLP, RF, and MLP all perform
well, with XGB lagging. We also generally observe qualitative consistency
as, for example, across all models the periodic scenario yields a large improvement in accuracy
for (\Malap, \Agent) and (\SMSreg, \Malap), while (\Airpush, \Boxer) and (\Boxer, \SMSreg)
only improve marginally. 
Furthermore, SVM achieves the highest accuracy in the static scenario 
for~40\%\ of the family combinations (8 out of~20),
followed by MLP and RF at 25\%\ each, with XGB performing best in 
the remaining~10\%\ of the combinations.
However, in the periodic retraining scenario, 
MLP performs best for~65\% of the family combinations, 
followed by SVM and RF at~25\%\ and~10\%, respectively, with XGB being best for
only~1 such combination.

\begin{figure}[!htb]
    \centering
    \begin{tikzpicture}[scale=0.9, every node/.style={scale=1.0}]
\pgfkeys{/pgf/number format/.cd,1000 sep={}}
\begin{axis}[
        width  = 0.65*\textwidth,
        height = 7.0cm,
        ymin=0.0,ymax=1.25,
        ytick={0.0, 0.2, 0.4, 0.6, 0.8, 1.0},
        major x tick style = transparent,
        ybar=5*\pgflinewidth,
        bar width=22.5pt,
        ylabel = {Average accuracy},
        ylabel style = {scale = 0.95},
        symbolic x coords={A, B, C, D},
        xticklabels={MLP, RF, SVM, XGB},
	y tick label style={scale=1.0,
    		/pgf/number format/.cd,
   		fixed,
   		fixed zerofill,
    		precision=2},
        xtick = data,
        x tick label style={scale=1.0,
		},
        nodes near coords,
        every node near coord/.append style={rotate=90, scale=0.85,
        								   anchor=west, 
								   /pgf/number format/.cd,
								   fixed,
								   fixed zerofill,
								   precision=4},
        enlarge x limits=0.165,
        legend cell align=left,
        legend pos=south east,
]
\addplot [fill=blue,opacity=1.00]
coordinates {
(A, 0.8248)
(B, 0.8152)
(C, 0.8416)
(D, 0.7578)
};
\addlegendentry{Static}
\addplot [fill=red,opacity=1.00]
coordinates {
(A, 0.9666)
(B, 0.9529)
(C, 0.9500)
(D, 0.9334)
};
\addlegendentry{Periodic}
\end{axis}
\end{tikzpicture}
    \caption{Average static and periodic accuracy per learning model}
    \label{fig:avgPeriodicStaticPerModel}
\end{figure}
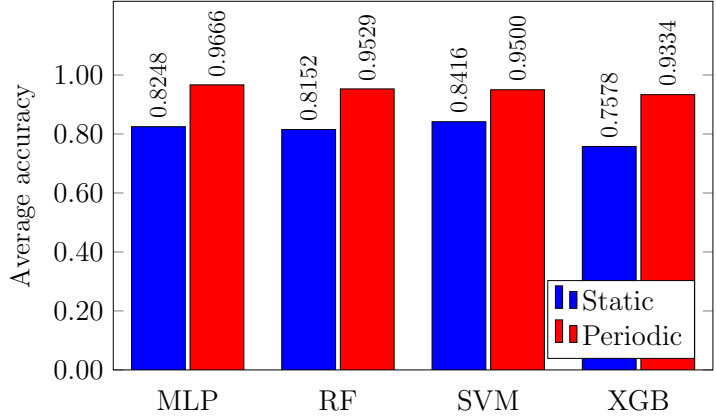

In Figure~\ref{fig:avgPeriodicStaticPerModel} we provide the
average~$A_{\static}$ and~$A_{\periodic}$ for each learning model across all family combinations.
This graph clearly demonstrates the value of accounting for concept drift, as the 
best average accuracy in the static scenario is~0.8416 (SVM model), while the worst
average accuracy in the periodic retraining scenario is~0.9334 (XGB model).
Furthermore, using the best model in the periodic scenario (MLP), we obtain an
average accuracy of~0.9666.

\begin{figure}[!htb]
    \centering
    \includegraphics[width=0.85\textwidth]{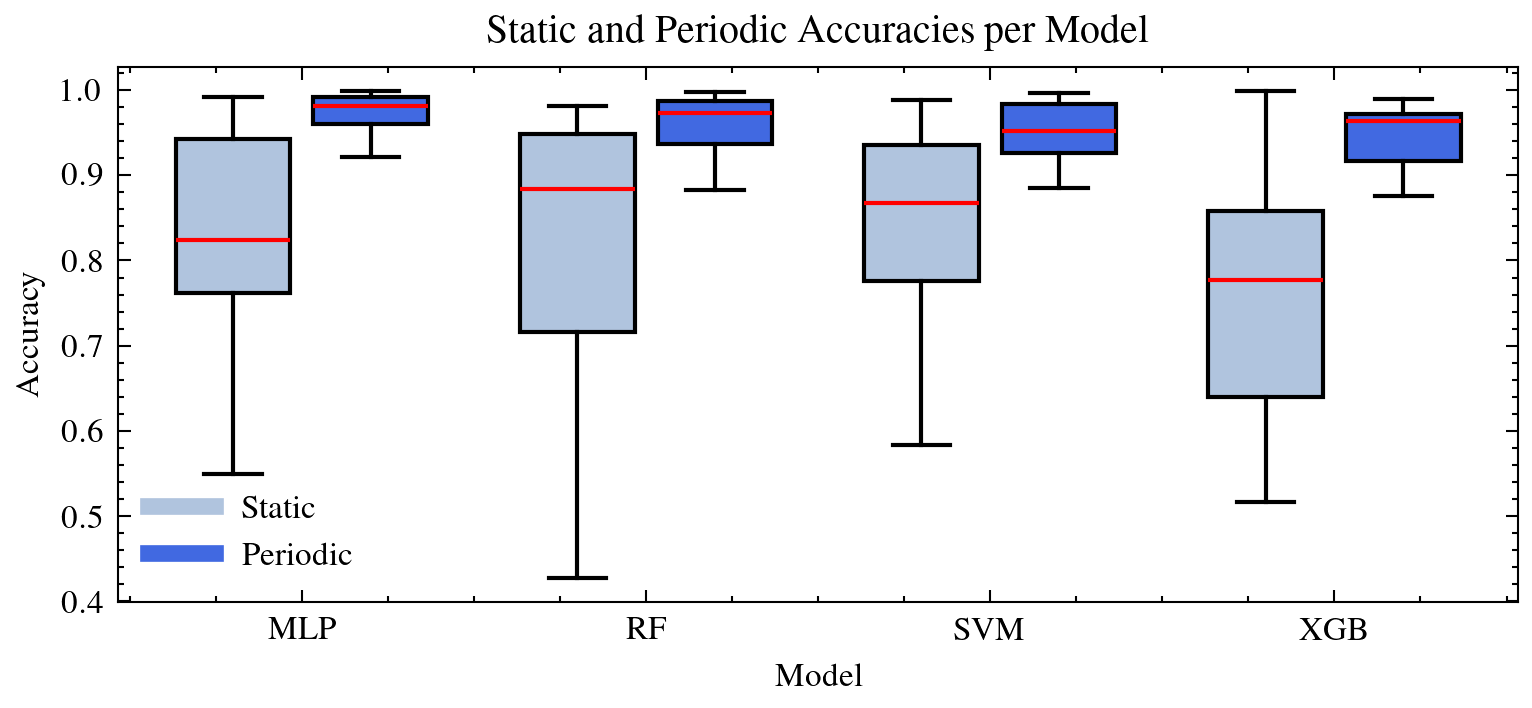}
    \caption{Box plot of static and periodic accuracy}
    \label{fig:periodicStaticPerModelBoxPlot}
\end{figure}

Another aspect to consider is the effect of hyperparameter tunning of each learning 
model instance.
If hyperparameter tunning was generally successful for learning model~$L$, 
we expect the distribution of its periodic accuracies to be more 
compact and have a higher median, as compared to the distribution of the static accuracies.
This is indeed the case, as illustrated in Figure~\ref{fig:periodicStaticPerModelBoxPlot}, 
which validates our use of TPE for hyperparameter tuning.

Overall, these static and periodic results confirm the success of
our experimental design and hyperparameter tuning. 
It is also clear that retraining models is necessary to 
counter concept drift. However, given the computational expense of training learning models, 
we would like to achieve comparable results to the periodic scenario with less work.
To this end, we now consider drift-aware retraining for each of the
three drift detectors introduced in Section~\ref{sect:driftTechniques}. Our primary goal is
to determine whether we can achieve results comparable to periodic retraining, 
but at a significantly lower cost, in terms of the number of models trained.

\subsection{MK-Means Drift-Aware Results}\label{sec:mkmeansDiscussion}

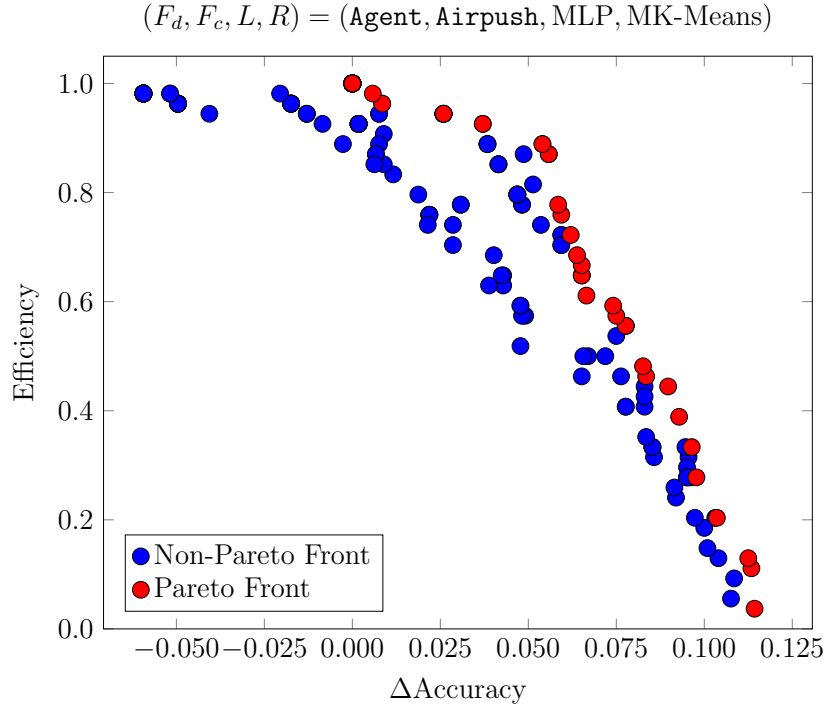
\begin{figure}[!htb]
    \centering
    \begin{tikzpicture}[scale=0.9, every node/.style={scale=1.0}]
\pgfkeys{/pgf/number format/.cd,1000 sep={}}
\begin{axis}[
	only marks,
        width  = 12.0cm,
        height = 10.0cm,
        ymin=0.0, ymax=1.05,
        ytick={0.0, 0.2, 0.4, 0.6, 0.8, 1.0},
        xmin=-0.065, xmax=0.125,
        xtick={-0.050, -0.025, 0.000, 0.025, 0.050, 0.075, 0.100, 0.125},
        xlabel = {$\Delta\mbox{Accuracy}$},
        xlabel style = {scale = 1.0},
        ylabel = {Efficiency},
        ylabel style = {scale = 1.0},
        title = {$(F_{d},F_{c},L,R)=(\mbox{\Agent},\mbox{\Airpush},\mlp,\mbox{MK-Means})$},
        title style = {scale = 1.0},
	x tick label style={scale=1.0,
    		/pgf/number format/.cd,
   		fixed,
   		fixed zerofill,
    		precision=3},
	y tick label style={scale=1.0,
    		/pgf/number format/.cd,
   		fixed,
   		fixed zerofill,
    		precision=1},
        enlarge x limits=0.03,
        legend cell align=left,
        legend pos=south west,
        legend style={nodes={scale=1.0},
        },
]
\addplot [fill=blue,opacity=1.00,mark=*,mark size=3.5pt]
coordinates {
(0.108482143,0.092592593)
(0.107589286,0.055555556)
(0.104017857,0.12962963)
(0.103125,0.203703704)
(0.100892857,0.148148148)
(0.1,0.185185185)
(0.097321429,0.203703704)
(0.095982143,0.277777778)
(0.095535714,0.314814815)
(0.095089286,0.296296296)
(0.095089286,0.277777778)
(0.095089286,0.277777778)
(0.094642857,0.333333333)
(0.091964286,0.240740741)
(0.091517857,0.259259259)
(0.085714286,0.314814815)
(0.085267857,0.333333333)
(0.085267857,0.333333333)
(0.085267857,0.333333333)
(0.083482143,0.351851852)
(0.083035714,0.444444444)
(0.083035714,0.407407407)
(0.083035714,0.425925926)
(0.077678571,0.555555556)
(0.077678571,0.555555556)
(0.077678571,0.407407407)
(0.077678571,0.407407407)
(0.076339286,0.462962963)
(0.075,0.537037037)
(0.071875,0.5)
(0.066964286,0.5)
(0.065625,0.5)
(0.065178571,0.648148148)
(0.065178571,0.648148148)
(0.065178571,0.462962963)
(0.059375,0.722222222)
(0.059375,0.722222222)
(0.059375,0.703703704)
(0.059375,0.703703704)
(0.055803571,0.87037037)
(0.054017857,0.888888889)
(0.053571429,0.740740741)
(0.051339286,0.814814815)
(0.049107143,0.574074074)
(0.048660714,0.87037037)
(0.048214286,0.777777778)
(0.048214286,0.777777778)
(0.048214286,0.574074074)
(0.047767857,0.518518519)
(0.047767857,0.592592593)
(0.047767857,0.592592593)
(0.046875,0.796296296)
(0.046875,0.796296296)
(0.046875,0.796296296)
(0.046875,0.796296296)
(0.046875,0.796296296)
(0.042857143,0.648148148)
(0.042857143,0.62962963)
(0.042410714,0.648148148)
(0.041517857,0.851851852)
(0.041517857,0.851851852)
(0.041517857,0.851851852)
(0.040178571,0.685185185)
(0.038839286,0.62962963)
(0.038392857,0.888888889)
(0.038392857,0.888888889)
(0.038392857,0.888888889)
(0.037053571,0.925925926)
(0.030803571,0.777777778)
(0.030803571,0.777777778)
(0.028571429,0.703703704)
(0.028571429,0.740740741)
(0.025892857,0.944444444)
(0.025892857,0.944444444)
(0.025892857,0.944444444)
(0.025892857,0.944444444)
(0.025892857,0.944444444)
(0.025892857,0.944444444)
(0.021875,0.759259259)
(0.021875,0.759259259)
(0.021875,0.759259259)
(0.021875,0.759259259)
(0.021428571,0.740740741)
(0.01875,0.796296296)
(0.011607143,0.833333333)
(0.008928571,0.907407407)
(0.008928571,0.851851852)
(0.008482143,0.962962963)
(0.007589286,0.944444444)
(0.007589286,0.944444444)
(0.007589286,0.888888889)
(0.006696429,0.87037037)
(0.006696429,0.87037037)
(0.00625,0.851851852)
(0.001785714,0.925925926)
(0.001785714,0.925925926)
(0.001785714,0.925925926)
(0.001785714,0.925925926)
(0.001785714,0.925925926)
(0.001785714,0.925925926)
(0.001785714,0.925925926)
(0.001785714,0.925925926)
(0.001785714,0.925925926)
(0.001785714,0.925925926)
(0,1)
(0,1)
(0,1)
(0,1)
(0,1)
(0,1)
(0,1)
(0,1)
(0,1)
(0,1)
(0,1)
(0,1)
(0,1)
(0,1)
(0,1)
(0,1)
(0,1)
(0,1)
(0,1)
(0,1)
(0,1)
(0,1)
(0,1)
(0,1)
(0,1)
(0,1)
(0,1)
(0,1)
(0,1)
(0,1)
(0,1)
(0,1)
(0,1)
(0,1)
(0,1)
(0,1)
(0,1)
(0,1)
(0,1)
(0,1)
(0,1)
(0,1)
(0,1)
(0,1)
(0,1)
(0,1)
(0,1)
(0,1)
(0,1)
(0,1)
(0,1)
(0,1)
(0,1)
(0,1)
(0,1)
(0,1)
(0,1)
(0,1)
(0,1)
(0,1)
(0,1)
(0,1)
(0,1)
(-0.002678571,0.888888889)
(-0.008482143,0.925925926)
(-0.012946429,0.944444444)
(-0.012946429,0.944444444)
(-0.012946429,0.944444444)
(-0.017410714,0.962962963)
(-0.017410714,0.962962963)
(-0.017410714,0.962962963)
(-0.017410714,0.962962963)
(-0.017410714,0.962962963)
(-0.017410714,0.962962963)
(-0.017410714,0.962962963)
(-0.017410714,0.962962963)
(-0.017410714,0.962962963)
(-0.020535714,0.981481481)
(-0.040625,0.944444444)
(-0.049553571,0.962962963)
(-0.049553571,0.962962963)
(-0.049553571,0.962962963)
(-0.049553571,0.962962963)
(-0.049553571,0.962962963)
(-0.049553571,0.962962963)
(-0.049553571,0.962962963)
(-0.049553571,0.962962963)
(-0.049553571,0.962962963)
(-0.049553571,0.962962963)
(-0.049553571,0.962962963)
(-0.049553571,0.962962963)
(-0.051785714,0.981481481)
(-0.051785714,0.981481481)
(-0.059375,0.981481481)
(-0.059375,0.981481481)
(-0.059375,0.981481481)
(-0.059375,0.981481481)
(-0.059375,0.981481481)
(-0.059375,0.981481481)
(-0.059375,0.981481481)
(-0.059375,0.981481481)
(-0.059375,0.981481481)
(-0.059375,0.981481481)
(-0.059375,0.981481481)
(-0.059375,0.981481481)
(-0.059375,0.981481481)
(-0.059375,0.981481481)
(-0.059375,0.981481481)
(-0.059375,0.981481481)
};
\addlegendentry{Non-Pareto Front}
\addplot [fill=red,opacity=1.00,mark=*,mark size=3.5pt]
coordinates {
(0.114285714,0.037037037)
(0.113392857,0.111111111)
(0.1125,0.12962963)
(0.103571429,0.203703704)
(0.097767857,0.277777778)
(0.096428571,0.333333333)
(0.092857143,0.388888889)
(0.089732143,0.444444444)
(0.083482143,0.462962963)
(0.082589286,0.481481481)
(0.077678571,0.555555556)
(0.075,0.574074074)
(0.074107143,0.592592593)
(0.066517857,0.611111111)
(0.065178571,0.648148148)
(0.065178571,0.666666667)
(0.063839286,0.685185185)
(0.062053571,0.722222222)
(0.059375,0.759259259)
(0.058482143,0.777777778)
(0.055803571,0.87037037)
(0.054017857,0.888888889)
(0.037053571,0.925925926)
(0.025892857,0.944444444)
(0.008482143,0.962962963)
(0.005803571,0.981481481)
(0,1)
};
\addlegendentry{Pareto Front}
\end{axis}
\end{tikzpicture}
    \caption{MK-Means Pareto Front example} 
    \label{fig:paretoFrontAgentAirpushMLP}
\end{figure}

Recall that MK-Means 
has two hyperparameters, $(\tau_{\mk}, \clusters)$. 
Figure~\ref{fig:paretoFrontAgentAirpushMLP} contains a plot of the 
Pareto Front for the specific case
$$
  (F_{d},F_{c},L,R)=(\mbox{\Agent},\mbox{\Airpush},\mlp,\mk)
$$
As previously noted, from the Pareto Front coordinates, it is trivial to determine the 
corresponding hyperparameters~$(\tau_{\mk}, \clusters)$.

From Figure~\ref{fig:paretoFrontAgentAirpushMLP} we can see the advantage of 
choosing hyperparameter values based on the Pareto Front. For example, if we require an
efficiency in the range of, say, 60\%\ to~65\%, choosing hyperparameter values
from the Pareto Front could provide an improvement in accuracy of more
than~3\%, as compared to values that are not on the Pareto Front.

\begin{figure}[!htb]
    \centering
    \begin{tikzpicture}[scale=0.9, every node/.style={scale=1.0}]
\pgfkeys{/pgf/number format/.cd,1000 sep={}}
\begin{axis}[
        width  = 0.8*\textwidth,
        height = 6.5cm,
        ymin=0.0,ymax=1.275,
        ytick={0.0, 0.2, 0.4, 0.6, 0.8, 1.0},
        major x tick style = transparent,
        ybar=5*\pgflinewidth,
        bar width=9.75pt,
        ylabel = {Accuracy},
        ylabel style = {scale=1.0},
        symbolic x coords={
        		A, B, C, D
        },
        xticklabels={
        		MLP, RF, SVM, XGB
        },
	y tick label style={scale=1.0,
    		/pgf/number format/.cd,
   		fixed,
   		fixed zerofill,
    		precision=2},
        xtick = data,
        x tick label style={scale=1.0,
		},
        nodes near coords,
        every node near coord/.append style={rotate=90, scale=0.825,
        								   anchor=west, 
								   /pgf/number format/.cd,
								   fixed,
								   fixed zerofill,
								   precision=4},
        enlarge x limits=0.175,
        legend cell align=left,
        legend pos=south east,
        legend columns=3,
        legend style={nodes={scale=0.9},
        },
]
\addplot [fill=blue,opacity=1.00]
coordinates {
(A, 0.8248) 
(B, 0.8152) 
(C, 0.8416) 
(D, 0.7578)
};
\addlegendentry{Static}
\addplot [fill=green,opacity=1.00]
coordinates {
(A, 0.886180076)
(B, 0.86860355)
(C, 0.881177378)
(D, 0.818501016)
};
\addlegendentry{\hbox{${\cal A}_{\drift}$}}
\addplot [fill=orange,opacity=1.00]
coordinates {
(A, 0.913474855)
(B, 0.894080535)
(C, 0.905159153)
(D, 0.852852292)
};
\addlegendentry{\hbox{${\cal A}^{x\kern-1pt\fame}_{\drift}$}}
\addplot [fill=yellow,opacity=1.00]
coordinates {
(A, 0.937658213)
(B, 0.923541877)
(C, 0.926644467)
(D, 0.887050375)
};
\addlegendentry{\hbox{${\cal A}^{y\kern-1pt\fame}_{\drift}$}}
\addplot [fill=cyan,opacity=1.00]
coordinates {
(A, 0.948011308)
(B, 0.935822808)
(C, 0.93778483)
(D, 0.902296157)
};
\addlegendentry{\hbox{${\cal A}^{m\kern-1pt\fame}_{\drift}$}}
\addplot [fill=red,opacity=1.00]
coordinates {
(A, 0.9666) 
(B, 0.9529) 
(C, 0.9500) 
(D, 0.9334)
};
\addlegendentry{Periodic}
\end{axis}
\end{tikzpicture}
    \caption{MK-Means accuracies}
    \label{fig:mkmeansFinalAccs}
\end{figure}
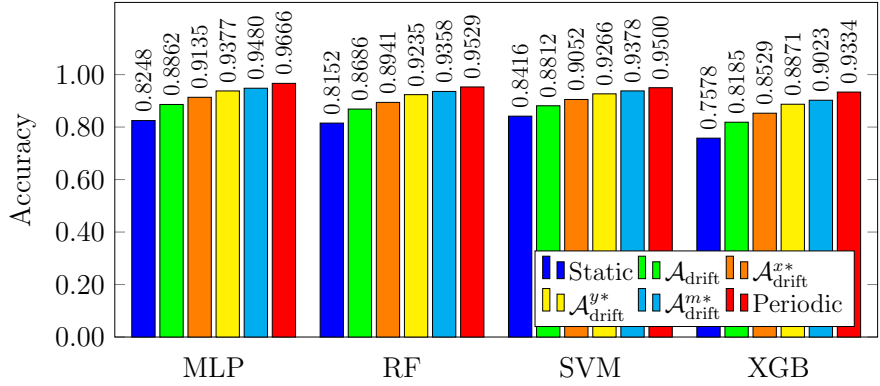

In Figure~\ref{fig:mkmeansFinalAccs}, we provide a bar graph 
comparing the average accuracies for each learning
model under the static, drift-aware, and periodic scenarios, where the
drift detection is based on MK-Means.
As discussed in Section~\ref{sec:consolidatingDriftExperiments},
for the drift-aware scenario,
we have four distinct methods for computing the average 
accuracy, which we denote as~${\cal A}_{\drift}$,
${\cal A}^{x\kern-1pt\fame}_{\drift}$, ${\cal A}^{y\kern-1pt\fame}_{\drift}$,
and~${\cal A}^{m\kern-1pt\fame}_{\drift}$. Recall that~${\cal A}_{\drift}$ is the
average accuracy over all~$(\tau_{\mk}, \clusters)$ pairs tested, while
the latter three averages are based on Pareto Front analysis.

From Figure~\ref{fig:mkmeansFinalAccs},
we observe that for all learning models tested,
using MK-Means for concept drift detection produces a gain
in accuracy, as compared to the static scenario. Also, by most of the measures
considered, drift-aware retraining based on MK-Means results in only a relatively small 
loss in accuracy as compared to the more costly periodic scenario.
For example, considering the median~${\cal A}^{m\kern-1pt\fame}_{\drift}$
for the MLP model, we see that the drift-aware scenario 
improves on the static scenario by nearly~15\%, since~$(0.9480-0.8248)/0.8248 = 0.1494$,
while being within~2\%\ of the accuracy achieved in the more costly periodic scenario,
since~$(0.9666-0.9480)/0.9666 = 0.0192$.

\begin{figure}[!htb]
    \centering
    \begin{tikzpicture}[scale=0.9, every node/.style={scale=1.0}]
\pgfkeys{/pgf/number format/.cd,1000 sep={}}
\begin{axis}[
        width  = 0.6*\textwidth,
        height = 6.5cm,
        ymin=0.0,ymax=1.125,
        ytick={0.0, 0.2, 0.4, 0.6, 0.8, 1.0},
        major x tick style = transparent,
        ybar=5*\pgflinewidth,
        bar width=9.75pt,
        ylabel = {Efficiency},
        ylabel style = {scale=1.0},
        symbolic x coords={
        		A, B, C, D
        },
        xticklabels={
        		MLP, RF, SVM, XGB
        },
	y tick label style={scale=1.0,
    		/pgf/number format/.cd,
   		fixed,
   		fixed zerofill,
    		precision=2},
        xtick = data,
        x tick label style={scale=1.0,
		},
        nodes near coords,
        every node near coord/.append style={rotate=90, scale=0.825,
        								   anchor=west, 
								   /pgf/number format/.cd,
								   fixed,
								   fixed zerofill,
								   precision=4},
        enlarge x limits=0.175,
        legend cell align=left,
        legend pos=south east,
        legend columns=2,
        legend style={nodes={scale=0.9},
        },
]
\addplot [fill=green,opacity=1.00]
coordinates {
(A, 0.820955391)
(B, 0.820955391)
(C, 0.820955391)
(D, 0.820955391)
};
\addlegendentry{\hbox{${\cal E}_{\drift}$}}
\addplot [fill=orange,opacity=1.00]
coordinates {
(A, 0.709300959)
(B, 0.74985382)
(C, 0.69891575)
(D, 0.704633472)
};
\addlegendentry{\hbox{${\cal E}^{x\kern-1pt\fame}_{\drift}$}}
\addplot [fill=yellow,opacity=1.00]
coordinates {
(A, 0.594866414)
(B, 0.704535685)
(C, 0.6707246)
(D, 0.667945781)
};
\addlegendentry{\hbox{${\cal E}^{y\kern-1pt\fame}_{\drift}$}}
\addplot [fill=cyan,opacity=1.00]
coordinates {
(A, 0.58669897)
(B, 0.748531829)
(C, 0.68991958)
(D, 0.696381591)
};
\addlegendentry{\hbox{${\cal E}^{m\kern-1pt\fame}_{\drift}$}}
\end{axis}
\end{tikzpicture}
    \caption{MK-Means efficiencies}
    \label{fig:mkmeansFinalEffs}
\end{figure}
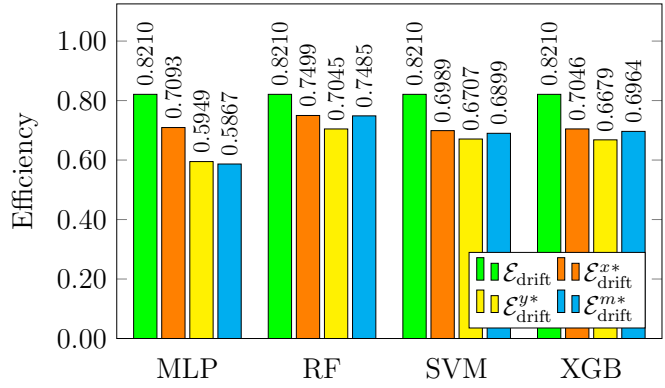

Figure~\ref{fig:mkmeansFinalEffs} shows the analogous efficiency
results. 
The improvement in efficiency for the drift-aware scenario
averaged over all hyperparameters~$(\tau_{\mk}, \clusters)$ 
tested, as indicated by~${\cal E}_{\drift}$, is over~80\%, although
the improvement when restricted to the
better choices of hyperparameters, as represented by
the Pareto Front-based values of~${\cal E}^{x\kern-1pt\fame}_{\drift}$, 
${\cal E}^{y\kern-1pt\fame}_{\drift}$, and~${\cal E}^{m\kern-1pt\fame}_{\drift}$
is not quite as impressive.
Nevertheless, for
the median~${\cal E}^{m\kern-1pt\fame}_{\drift}$ we achieve a 
savings---in terms of the number of models that must be trained---of 
more than~58\%, as compared to the periodic retraining scenario.

\subsection{OCSVM Drift-Aware Results}\label{sec:ocsvmDiscussion}

\begin{figure}[!htb]
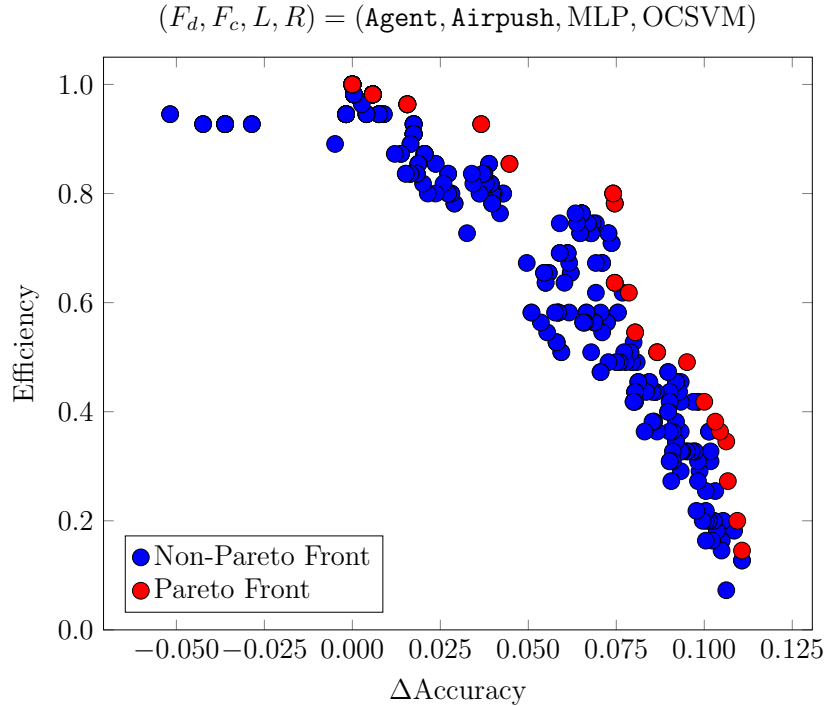

    \centering
    \input figures/pareto_aa_mlp_oc.tex
    \caption{OCSVM Pareto Front example} 
    \label{fig:ocsvmMalapAirpushMlpParetoFront}
\end{figure}

Our analysis of concept drift detection based on OCSVM 
is similar to that for MK-Means, above.
In Figure~\ref{fig:ocsvmMalapAirpushMlpParetoFront}, we give the Pareto Front for the
same specific case that we considered for MK-Means, namely,
$$
  (F_{d},F_{c},L,R)=(\mbox{\Agent},\mbox{\Airpush},\mlp,\oc)
$$
As in the MK-Means case, we again see the value of Pareto Front analysis 
for selecting hyperparameters to meet a desired balance between accuracy and efficiency.

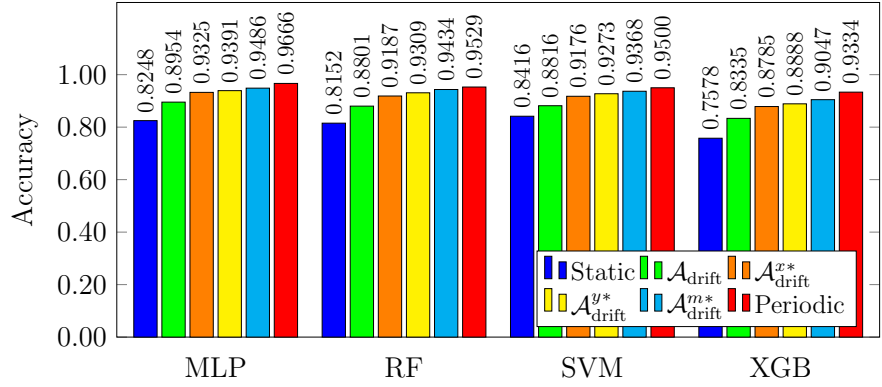
\begin{figure}[!htb]
    \centering
    \begin{tikzpicture}[scale=0.9, every node/.style={scale=1.0}]
\pgfkeys{/pgf/number format/.cd,1000 sep={}}
\begin{axis}[
        width  = 0.8*\textwidth,
        height = 6.5cm,
        ymin=0.0,ymax=1.275,
        ytick={0.0, 0.2, 0.4, 0.6, 0.8, 1.0},
        major x tick style = transparent,
        ybar=5*\pgflinewidth,
        bar width=9.75pt,
        ylabel = {Accuracy},
        ylabel style = {scale=1.0},
        symbolic x coords={
        		A, B, C, D
        },
        xticklabels={
        		MLP, RF, SVM, XGB
        },
	y tick label style={scale=1.0,
    		/pgf/number format/.cd,
   		fixed,
   		fixed zerofill,
    		precision=2},
        xtick = data,
        x tick label style={scale=1.0,
		},
        nodes near coords,
        every node near coord/.append style={rotate=90, scale=0.825,
        								   anchor=west, 
								   /pgf/number format/.cd,
								   fixed,
								   fixed zerofill,
								   precision=4},
        enlarge x limits=0.175,
        legend cell align=left,
        legend pos=south east,
        legend columns=3,
        legend style={nodes={scale=0.9},
        },
]
\addplot [fill=blue,opacity=1.00]
coordinates {
(A, 0.8248) 
(B, 0.8152) 
(C, 0.8416) 
(D, 0.7578)
};
\addlegendentry{Static}
\addplot [fill=green,opacity=1.00]
coordinates {
(A, 0.895353431)
(B, 0.880099774)
(C, 0.881618952)
(D, 0.833490096)
};
\addlegendentry{\hbox{${\cal A}_{\drift}$}}
\addplot [fill=orange,opacity=1.00]
coordinates {
(A, 0.932478949)
(B, 0.918660319)
(C, 0.917635083)
(D, 0.878527927)
};
\addlegendentry{\hbox{${\cal A}^{x\kern-1pt\fame}_{\drift}$}}
\addplot [fill=yellow,opacity=1.00]
coordinates {
(A, 0.939087443)
(B, 0.930871602)
(C, 0.92728501)
(D, 0.888842459)
};
\addlegendentry{\hbox{${\cal A}^{y\kern-1pt\fame}_{\drift}$}}
\addplot [fill=cyan,opacity=1.00]
coordinates {
(A, 0.948600043)
(B, 0.943418163)
(C, 0.93678848)
(D, 0.904678914)
};
\addlegendentry{\hbox{${\cal A}^{m\kern-1pt\fame}_{\drift}$}}
\addplot [fill=red,opacity=1.00]
coordinates {
(A, 0.9666) 
(B, 0.9529) 
(C, 0.9500) 
(D, 0.9334)
};
\addlegendentry{Periodic}
\end{axis}
\end{tikzpicture}
    \caption{OCSMV accuracies}
    \label{fig:ocsvmFinalAccs}
\end{figure}

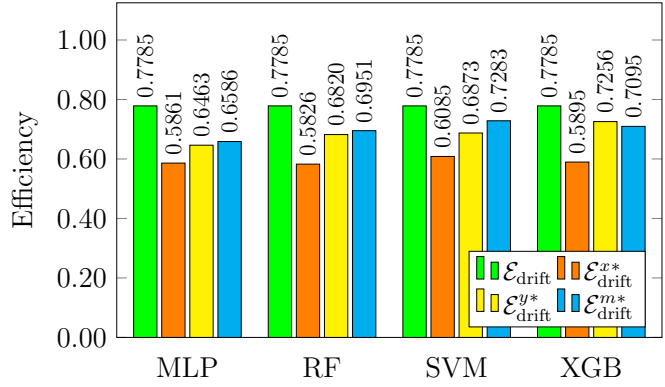
\begin{figure}[!htb]
    \centering
    \begin{tikzpicture}[scale=0.9, every node/.style={scale=1.0}]
\pgfkeys{/pgf/number format/.cd,1000 sep={}}
\begin{axis}[
        width  = 0.6*\textwidth,
        height = 6.5cm,
        ymin=0.0,ymax=1.125,
        ytick={0.0, 0.2, 0.4, 0.6, 0.8, 1.0},
        major x tick style = transparent,
        ybar=5*\pgflinewidth,
        bar width=9.75pt,
        ylabel = {Efficiency},
        ylabel style = {scale=1.0},
        symbolic x coords={
        		A, B, C, D
        },
        xticklabels={
        		MLP, RF, SVM, XGB
        },
	y tick label style={scale=1.0,
    		/pgf/number format/.cd,
   		fixed,
   		fixed zerofill,
    		precision=2},
        xtick = data,
        x tick label style={scale=1.0,
		},
        nodes near coords,
        every node near coord/.append style={rotate=90, scale=0.825,
        								   anchor=west, 
								   /pgf/number format/.cd,
								   fixed,
								   fixed zerofill,
								   precision=4},
        enlarge x limits=0.175,
        legend cell align=left,
        legend pos=south east,
        legend columns=2,
        legend style={nodes={scale=0.9},
        },
]
\addplot [fill=green,opacity=1.00]
coordinates {
(A, 0.778471942)
(B, 0.778471942)
(C, 0.778471942)
(D, 0.778471942)
};
\addlegendentry{\hbox{${\cal E}_{\drift}$}}
\addplot [fill=orange,opacity=1.00]
coordinates {
(A, 0.586140749)
(B, 0.582642896)
(C, 0.608460957)
(D, 0.589543811)
};
\addlegendentry{\hbox{${\cal E}^{x\kern-1pt\fame}_{\drift}$}}
\addplot [fill=yellow,opacity=1.00]
coordinates {
(A, 0.646267703)
(B, 0.682008949)
(C, 0.687274951)
(D, 0.725605201)
};
\addlegendentry{\hbox{${\cal E}^{y\kern-1pt\fame}_{\drift}$}}
\addplot [fill=cyan,opacity=1.00]
coordinates {
(A, 0.658601841)
(B, 0.695055424)
(C, 0.728337521)
(D, 0.709549682)
};
\addlegendentry{\hbox{${\cal E}^{m\kern-1pt\fame}_{\drift}$}}
\end{axis}
\end{tikzpicture}
    \caption{OCSVM efficiencies}
    \label{fig:ocsvmFinalEffs}
\end{figure}

In Figures~\ref{fig:ocsvmFinalAccs} and~\ref{fig:ocsvmFinalEffs}, respecitvely,
we provide bar graphs of our various accuracy and efficiency metrics.
We note that, similar to MK-Means, OCSVM achieves a substantial accuracy gain over the static baseline 
at a significant lower retraining cost than the periodic retraining scenario.
For example, if we consider the median~${\cal A}^{m\kern-1pt\fame}_{\drift}$
for the MLP model, the drift-aware scenario 
improves on the static scenario by~15\% ($(0.9486-0.8248)/0.8248 = 0.1501$),
while being within~1.9\%\ of the accuracy achieved in the more costly periodic scenario
($(0.9666-0.9486)/0.9666 = 0.0186$), a marginal improvement
over MK-Means.

With respect to efficiency, the improvement for the OCSVM-based 
drift-aware scenario averaged over all hyperparameters 
tested is more than~77\%. 
The improvement in efficiency when restricted to 
the median Pareto Front-based 
value~${\cal E}^{m\kern-1pt\fame}_{\drift}$ and MLP model
is more than~65\%. Recall that the corresponding median value
MK-Means is about~58\%. Thus, by this measure, OCSVM
offers improved efficiency, as compared to MK-Means.

\subsection{MMD Drift-Aware Results}\label{sec:mmdDiscussion}

Our MMD-based concept drift detection technique differs
from MK-Means and OCSVM in that for MMD we only consider one 
hyperparameter. This hyperparameter, which we denote as~$\tau_{\mmds}$,
is a confidence value for hypothesis testing.

\begin{figure}[!htb]
    \centering
    \begin{tikzpicture}[scale=0.9, every node/.style={scale=1.0}]
\pgfkeys{/pgf/number format/.cd,1000 sep={}}
\begin{axis}[
	only marks,
        width  = 12.0cm,
        height = 10.0cm,
        ymin=0.70, ymax=0.88,
        ytick={0.75, 0.80, 0.85},
        xmin=-0.004, xmax=0.038,
        xtick={0.00, 0.01, 0.02, 0.03},
        xlabel = {$\Delta\mbox{Accuracy}$},
        xlabel style = {scale = 1.0},
        ylabel = {Efficiency},
        ylabel style = {scale = 1.0},
        title = {$(F_{\drift},F_{\control},L,R)=(\mbox{\Agent},\mbox{\Airpush},\mlp,\mbox{MMD})$},
        title style = {scale = 1.0},
	x tick label style={scale=1.0,
    		/pgf/number format/.cd,
   		fixed,
   		fixed zerofill,
    		precision=2},
	y tick label style={scale=1.0,
    		/pgf/number format/.cd,
   		fixed,
   		fixed zerofill,
    		precision=2},
        scaled x ticks = false,
        legend cell align=left,
        legend pos=south west,
        legend style={nodes={scale=1.0},
        },
]
\addplot [fill=blue,opacity=1.00,mark=*,mark size=3.5pt]
coordinates {
(0.035267857,0.709090909)
(0.035267857,0.709090909)
(0.026785714,0.709090909)
(0.026785714,0.709090909)
(0.026785714,0.709090909)
(0.025892857,0.727272727)
(0.025892857,0.727272727)
(0.025892857,0.727272727)
(0.025892857,0.727272727)
(0.025892857,0.727272727)
(0.025892857,0.727272727)
(0.025892857,0.727272727)
(0.025892857,0.727272727)
(0.025892857,0.727272727)
(0.025892857,0.727272727)
(0.025892857,0.727272727)
(0.025892857,0.727272727)
(0.025892857,0.727272727)
(0.025892857,0.727272727)
(0.025892857,0.727272727)
(0.025892857,0.727272727)
(0.025892857,0.727272727)
(0.025892857,0.727272727)
(0.025892857,0.727272727)
(0.025892857,0.727272727)
(0.025892857,0.727272727)
(0.025892857,0.727272727)
(0.025892857,0.727272727)
(0.025892857,0.727272727)
(0.025892857,0.727272727)
(0.025892857,0.727272727)
(0.025892857,0.727272727)
(0.025892857,0.727272727)
(0.025892857,0.727272727)
(0.025892857,0.727272727)
(0.025892857,0.727272727)
(0.025892857,0.727272727)
(0.025892857,0.727272727)
(0.025892857,0.727272727)
(0.025892857,0.727272727)
(0.025892857,0.727272727)
(0.025892857,0.727272727)
(0.025892857,0.727272727)
(0.025892857,0.727272727)
(0.025892857,0.727272727)
(0.025892857,0.727272727)
(0.025892857,0.727272727)
(0.025892857,0.727272727)
(0.025892857,0.727272727)
(0.025892857,0.727272727)
(0.025892857,0.727272727)
(0.025892857,0.727272727)
(0.025892857,0.727272727)
(0.025892857,0.727272727)
(0.025892857,0.727272727)
(0.025892857,0.727272727)
(0.025892857,0.727272727)
(0.025892857,0.727272727)
(0.025892857,0.727272727)
(0.025892857,0.727272727)
(0.025892857,0.727272727)
(0.025892857,0.727272727)
(0.025892857,0.727272727)
(0.025892857,0.727272727)
(0.025892857,0.727272727)
(0.025,0.745454545)
(0.025,0.745454545)
(0.025,0.745454545)
(0.024553571,0.745454545)
(0.024553571,0.745454545)
(0.024107143,0.745454545)
(0.024107143,0.745454545)
(0.024107143,0.745454545)
(0.024107143,0.745454545)
(0.024107143,0.745454545)
(0.023660714,0.763636364)
(0.023214286,0.763636364)
(0.023214286,0.763636364)
(0.022767857,0.763636364)
(0.022767857,0.763636364)
(0.021428571,0.781818182)
(0.021428571,0.781818182)
(0.020535714,0.8)
(0.020089286,0.8)
(0.01875,0.8)
(0.018303571,0.818181818)
(0.009821429,0.836363636)
(0.009821429,0.836363636)
(0.008482143,0.854545455)
(0.0,0.8)
(-0.002678571,0.836363636)
};
\addlegendentry{Non-Pareto Front}
\addplot [fill=red,opacity=1.00,mark=*,mark size=3.5pt]
coordinates {
(0.035267857,0.709090909)
(0.025892857,0.727272727)
(0.025446429,0.745454545)
(0.024553571,0.763636364)
(0.021875,0.781818182)
(0.020535714,0.818181818)
(0.01875,0.836363636)
(0.009375,0.854545455)
(0.003571429,0.872727273)
};
\addlegendentry{Pareto Front}
\end{axis}
\end{tikzpicture}
    \caption{MMD Pareto Front example} 
    \label{fig:mmdMalapAirpushParetoFront}
\end{figure}
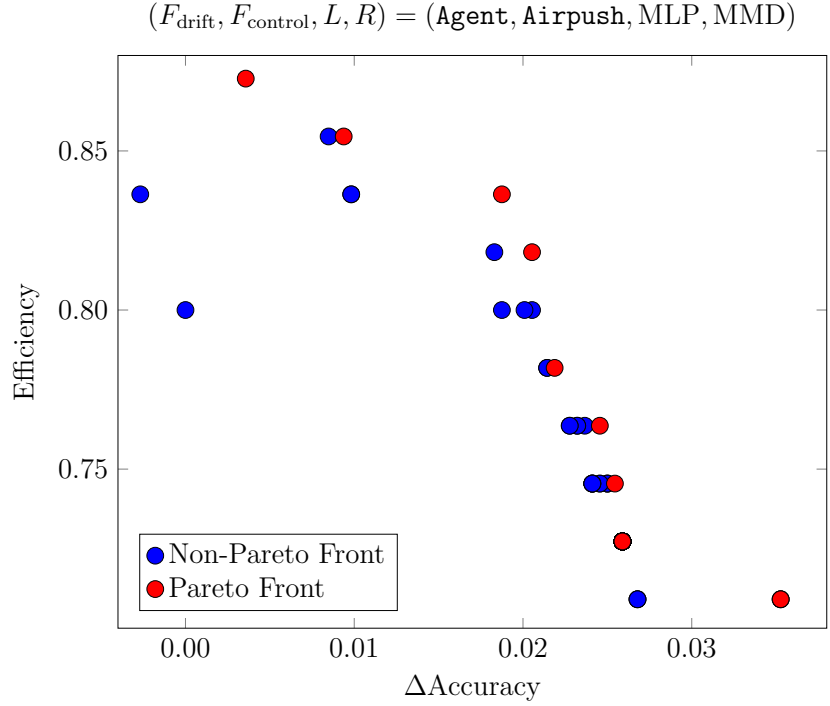

In Figure~\ref{fig:mmdMalapAirpushParetoFront}, we give the Pareto Front for the
same specific case that we considered for both MK-Means and OCSVM, that is,
$$
  (F_{d},F_{c},L,R)=(\mbox{\Agent},\mbox{\Airpush},\mlp,\mmds)
$$
Compared to MK-Means and OCSVM, 
we observe that for this MMD example, there is more consistency in the sense that
points tend to be very close to the Pareto Front, modulo a small number of outliers.

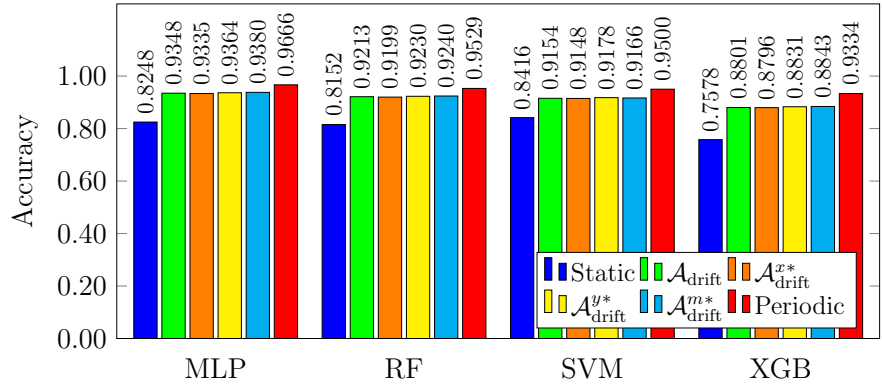
\begin{figure}[!htb]
    \centering
    \begin{tikzpicture}[scale=0.9, every node/.style={scale=1.0}]
\pgfkeys{/pgf/number format/.cd,1000 sep={}}
\begin{axis}[
        width  = 0.8*\textwidth,
        height = 6.5cm,
        ymin=0.0,ymax=1.275,
        ytick={0.0, 0.2, 0.4, 0.6, 0.8, 1.0},
        major x tick style = transparent,
        ybar=5*\pgflinewidth,
        bar width=9.75pt,
        ylabel = {Accuracy},
        ylabel style = {scale=1.0},
        symbolic x coords={
        		A, B, C, D
        },
        xticklabels={
        		MLP, RF, SVM, XGB
        },
	y tick label style={scale=1.0,
    		/pgf/number format/.cd,
   		fixed,
   		fixed zerofill,
    		precision=2},
        xtick = data,
        x tick label style={scale=1.0,
		},
        nodes near coords,
        every node near coord/.append style={rotate=90, scale=0.825,
        								   anchor=west, 
								   /pgf/number format/.cd,
								   fixed,
								   fixed zerofill,
								   precision=4},
        enlarge x limits=0.175,
        legend cell align=left,
        legend pos=south east,
        legend columns=3,
        legend style={nodes={scale=0.9},
        },
]
\addplot [fill=blue,opacity=1.00]
coordinates {
(A, 0.8248) 
(B, 0.8152) 
(C, 0.8416) 
(D, 0.7578)
};
\addlegendentry{Static}
\addplot [fill=green,opacity=1.00]
coordinates {
(A, 0.934751675)
(B, 0.921349475)
(C, 0.915380025)
(D, 0.880098012)
};
\addlegendentry{\hbox{${\cal A}_{\drift}$}}
\addplot [fill=orange,opacity=1.00]
coordinates {
(A, 0.933516961)
(B, 0.919899613)
(C, 0.914828158)
(D, 0.879622486)
};
\addlegendentry{\hbox{${\cal A}^{x\kern-1pt\fame}_{\drift}$}}
\addplot [fill=yellow,opacity=1.00]
coordinates {
(A, 0.936428823)
(B, 0.922961532)
(C, 0.917784135)
(D, 0.883062607)
};
\addlegendentry{\hbox{${\cal A}^{y\kern-1pt\fame}_{\drift}$}}
\addplot [fill=cyan,opacity=1.00]
coordinates {
(A, 0.937966684)
(B, 0.924023534)
(C, 0.91657302)
(D, 0.884276155)
};
\addlegendentry{\hbox{${\cal A}^{m\kern-1pt\fame}_{\drift}$}}
\addplot [fill=red,opacity=1.00]
coordinates {
(A, 0.9666) 
(B, 0.9529) 
(C, 0.9500) 
(D, 0.9334)
};
\addlegendentry{Periodic}
\end{axis}
\end{tikzpicture}
    \caption{MMD accuracies}
    \label{fig:mmdFinalAccs}
\end{figure}

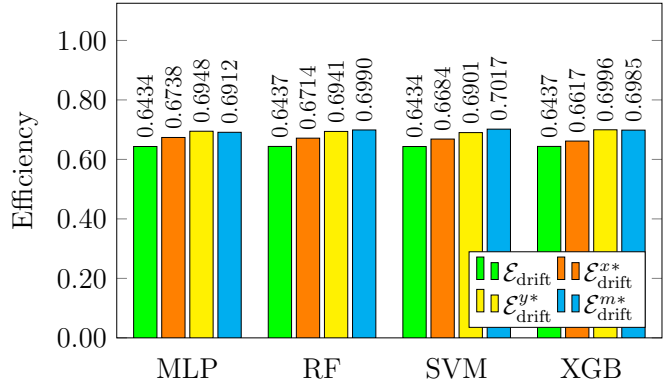
\begin{figure}[!htb]
    \centering
    \begin{tikzpicture}[scale=0.9, every node/.style={scale=1.0}]
\pgfkeys{/pgf/number format/.cd,1000 sep={}}
\begin{axis}[
        width  = 0.6*\textwidth,
        height = 6.5cm,
        ymin=0.0,ymax=1.125,
        ytick={0.0, 0.2, 0.4, 0.6, 0.8, 1.0},
        major x tick style = transparent,
        ybar=5*\pgflinewidth,
        bar width=9.75pt,
        ylabel = {Efficiency},
        ylabel style = {scale=1.0},
        symbolic x coords={
        		A, B, C, D
        },
        xticklabels={
        		MLP, RF, SVM, XGB
        },
	y tick label style={scale=1.0,
    		/pgf/number format/.cd,
   		fixed,
   		fixed zerofill,
    		precision=2},
        xtick = data,
        x tick label style={scale=1.0,
		},
        nodes near coords,
        every node near coord/.append style={rotate=90, scale=0.825,
        								   anchor=west, 
								   /pgf/number format/.cd,
								   fixed,
								   fixed zerofill,
								   precision=4},
        enlarge x limits=0.175,
        legend cell align=left,
        legend pos=south east,
        legend columns=2,
        legend style={nodes={scale=0.9},
        },
]
\addplot [fill=green,opacity=1.00]
coordinates {
(A, 0.643403917)
(B, 0.643665857)
(C, 0.643350926)
(D, 0.643706107)
};
\addlegendentry{\hbox{${\cal E}_{\drift}$}}
\addplot [fill=orange,opacity=1.00]
coordinates {
(A, 0.673761339)
(B, 0.671442051)
(C, 0.668423332)
(D, 0.6616706)
};
\addlegendentry{\hbox{${\cal E}^{x\kern-1pt\fame}_{\drift}$}}
\addplot [fill=yellow,opacity=1.00]
coordinates {
(A, 0.694772125)
(B, 0.694088616)
(C, 0.690128976)
(D, 0.699604999)
};
\addlegendentry{\hbox{${\cal E}^{y\kern-1pt\fame}_{\drift}$}}
\addplot [fill=cyan,opacity=1.00]
coordinates {
(A, 0.691172002)
(B, 0.69900472)
(C, 0.701739978)
(D, 0.698468801)
};
\addlegendentry{\hbox{${\cal E}^{m\kern-1pt\fame}_{\drift}$}}
\end{axis}
\end{tikzpicture}
    \caption{MMD efficiencies}
    \label{fig:mmdFinalEffs}
\end{figure}

In Figure~\ref{fig:mmdFinalAccs}  we provide bar graphs of the various accuracy metrics 
and in Figure~\ref{fig:mmdFinalEffs} we give the corresponding efficiency results.
As with both MK-Means and OCSVM, MMD achieves a substantial gain over the static baseline 
at a significant lower retraining cost, as compared with the periodic retraining scenario.
For example, if we consider the median~${\cal A}^{m\kern-1pt\fame}_{\drift}$
for the MLP model, the drift-aware scenario 
improves on the static scenario by~13.72\%, since~$(0.9380-0.8248)/0.8248 = 0.1372$,
while being within~3\%\ of the accuracy achieved in the much more costly periodic retraining scenario,
since~$(0.9666-0.9380)/0.9666 = 0.0296$. These numbers are slightly
worse than the results we obtained using MK-Means and OCSVM.

With respect to efficiency, the improvement over the periodic scenario,
averaged over all hyperparameters 
tested, is more than~64\%.  On the other hand,
the efficiency when restricted to Pareto Front
is slightly better---for example, the 
value~${\cal E}^{m\kern-1pt\fame}_{\drift}$ for the MLP
model is more than~69\%. Recall that the corresponding median value
MK-Means is about~58\%, and for OCSVM it is about~65\%.

For MK-Means and OCSVM, the average efficiency
declined significantly when we restricted the hyperparameters to
the Pareto Front. This indicates that for MK-Means and OCSVM,
many of the hyperparameter choices 
fail to adequately detect concept drift, while for the better 
hyperparameter choices (i.e., those on the Pareto Front), 
we obtain strong results. This is not the case for MMD, where
hyperparameter choices yield more consistent efficiency results,
regardless of whether we restrict to the Pareto Front.
The violin graphs in Figures~\ref{fig:violinAll} and~\ref{fig:violinPareto}
serve to further emphasize these points.
These results strongly suggest that we could reduce the hyperparameter
search space for MK-Means and OCSVM without affecting the 
accuracy or training efficiency results.

\subsection{Comparison of Drift Detection Techniques}

In this section we provide a comparative analysis of the three drift detection techniques
discussed above, with the emphasis on aggregate results. We pay particular 
attention to~${\cal A}^{m\kern-1pt\fame}_{\drift}$ and~${\cal E}^{m\kern-1pt\fame}_{\drift}$, 
since since these values correspond to typical selections of
drift-detector hyperparameters 

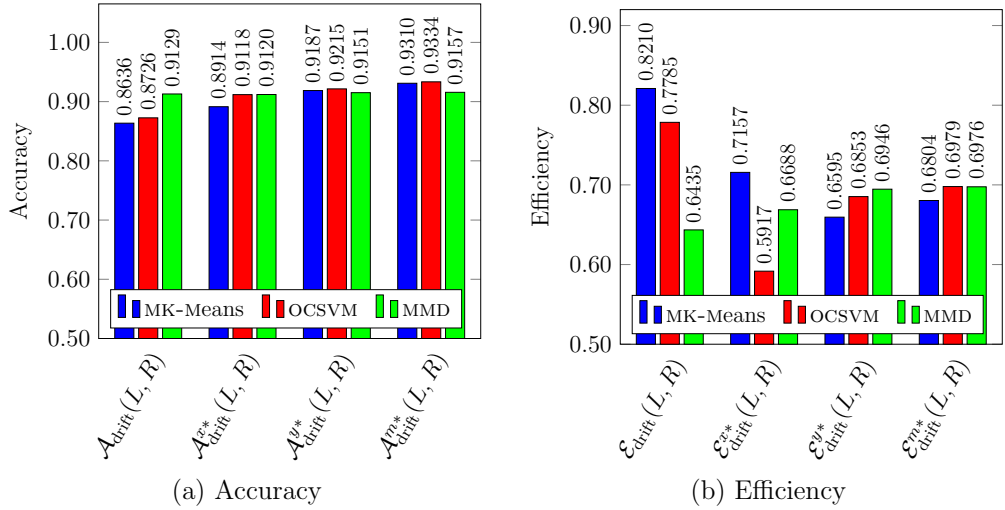
\begin{figure}[!htb]
    \centering
    \begin{tabular}{cc}
    \adjustbox{scale=1.0}{%
    \begin{tikzpicture}[scale=0.9, every node/.style={scale=1.0}]
\pgfkeys{/pgf/number format/.cd,1000 sep={}}
\begin{axis}[
        width  = 0.45*\textwidth,
        height = 6.5cm,
        ymin=0.5,ymax=1.065,
        ytick={0.5, 0.6, 0.7, 0.8, 0.9, 1.0},
        major x tick style = transparent,
        ybar=5*\pgflinewidth,
        bar width=8.0pt,
        ylabel = {Accuracy},
        ylabel style = {scale=0.85},
        symbolic x coords={
        		A, B, C, D
        },
        xticklabels={
       		\mbox{${\cal A}_{\drift}(L,R)$}, 
		\mbox{${\cal A}^{x\kern-1pt\fame}_{\drift}(L,R)$}, 
		\mbox{${\cal A}^{y\kern-1pt\fame}_{\drift}(L,R)$}, 
		\mbox{${\cal A}^{m\kern-1pt\fame}_{\drift}(L,R)$}
        },
	y tick label style={scale=0.85,
    		/pgf/number format/.cd,
   		fixed,
   		fixed zerofill,
    		precision=2},
        xtick = data,
        x tick label style={scale=0.85,
        		rotate=60,
		anchor=north east,
		inner sep=0mm
		},
        nodes near coords,
        every node near coord/.append style={rotate=90, scale=0.75,
        								   anchor=west, 
								   /pgf/number format/.cd,
								   fixed,
								   fixed zerofill,
								   precision=4},
        enlarge x limits=0.175,
        legend cell align=left,
        legend columns=3,
        legend pos=south west,
        legend style={nodes={scale=0.8},
        },
]
\addplot [fill=blue,opacity=1.00]
coordinates {
(A,0.8636) 
(B,0.8914) 
(C,0.9187) 
(D,0.9310)
};
\addlegendentry{\mbox{$\mk$}\hspace*{0.1in}}
\addplot [fill=red,opacity=1.00]
coordinates {
(A,0.8726) 
(B,0.9118) 
(C,0.9215) 
(D,0.9334)
};
\addlegendentry{\mbox{$\oc$}\hspace*{0.1in}}
\addplot [fill=green,opacity=1.00]
coordinates {
(A,0.9129) 
(B,0.9120) 
(C,0.9151) 
(D,0.9157)
};
\addlegendentry{\mbox{$\mmds$}}
\end{axis}
\end{tikzpicture}
    }
    &
    \adjustbox{scale=1.0}{%
    \begin{tikzpicture}[scale=0.9, every node/.style={scale=1.0}]
\pgfkeys{/pgf/number format/.cd,1000 sep={}}
\begin{axis}[
        width  = 0.45*\textwidth,
        height = 6.5cm,
        ymin=0.5,ymax=0.92,
        ytick={0.5, 0.6, 0.7, 0.8, 0.9},
        major x tick style = transparent,
        ybar=5*\pgflinewidth,
        bar width=8.0pt,
        ylabel = {Efficiency},
        ylabel style = {scale=0.85},
        symbolic x coords={
        		A, B, C, D
        },
        xticklabels={
       		\mbox{${\cal E}_{\drift}(L,R)$}, 
		\mbox{${\cal E}^{x\kern-1pt\fame}_{\drift}(L,R)$}, 
		\mbox{${\cal E}^{y\kern-1pt\fame}_{\drift}(L,R)$}, 
		\mbox{${\cal E}^{m\kern-1pt\fame}_{\drift}(L,R)$}
        },
	y tick label style={scale=0.85,
    		/pgf/number format/.cd,
   		fixed,
   		fixed zerofill,
    		precision=2},
        xtick = data,
        x tick label style={scale=0.85,
        		rotate=60,
		anchor=north east,
		inner sep=0mm
		},
        nodes near coords,
        every node near coord/.append style={rotate=90, scale=0.75,
        								   anchor=west, 
								   /pgf/number format/.cd,
								   fixed,
								   fixed zerofill,
								   precision=4},
        enlarge x limits=0.175,
        legend cell align=left,
        legend columns=3,
        legend pos=south west,
        legend style={nodes={scale=0.8},
        },
]
\addplot [fill=blue,opacity=1.00]
coordinates {
(A,0.8210) 
(B,0.7157) 
(C,0.6595) 
(D,0.6804)
};
\addlegendentry{\mbox{$\mk$}\hspace*{0.1in}}
\addplot [fill=red,opacity=1.00]
coordinates {
(A,0.7785) 
(B,0.5917) 
(C,0.6853) 
(D,0.6979)
};
\addlegendentry{\mbox{$\oc$}\hspace*{0.1in}}
\addplot [fill=green,opacity=1.00]
coordinates {
(A,0.6435) 
(B,0.6688) 
(C,0.6946) 
(D,0.6976)
};
\addlegendentry{\mbox{$\mmds$}}
\end{axis}
\end{tikzpicture}
    }
    \\[-0.25ex]
    \adjustbox{scale=0.85}{(a) Accuracy}
    &
    \adjustbox{scale=0.85}{(b) Efficiency}
     \end{tabular}
    \caption{Drift detector comparison averaged over all~$(F_{d},F{c},L)$}\label{fig:avg_acc_eff_perRL}
\end{figure}

In Figure~\ref{fig:avg_acc_eff_perRL}(a), we give accuracy results, averaged over
all~$(F_{d},F_{c},L)$, that is, all selections of drift family, control family,
and learning model. Figure~\ref{fig:avg_acc_eff_perRL}(b) gives the corresponding
efficiency averages.

From Figure~\ref{fig:avg_acc_eff_perRL} we see 
that the best~${\cal A}^{m\kern-1pt\fame}_{\drift}$ is obtained with the OCSVM
drift-detector, while MK-Means performs only marginally worse. 
With respect to~${\cal E}^{m\kern-1pt\fame}_{\drift}$, OCSVM and MMD 
are virtually identical.

Recall that we define efficiency in terms of the number of times that
concept drift is detected, which necessitates the retraining
of the learning models.
Since the same learning models are trained for all three drift detectors, the time required to
train any individual learning model has no bearing on this measure of 
efficiency. However, the time required to train each of our three 
drift detection techniques varies significantly, and this is relevant 
when selecting between these techniques.

\begin{figure}[!htb]
    \centering
    \begin{tikzpicture}[scale=0.9, every node/.style={scale=1.0}]
\pgfkeys{/pgf/number format/.cd,1000 sep={}}
\begin{axis}[
        width  = 0.6*\textwidth,
        height = 7.5cm,
        ymin=0.0,ymax=9.0,
        ytick={0.0, 2.0, 4.0, 6.0, 8.0},
        major x tick style = transparent,
        ybar=5*\pgflinewidth,
        bar width=14.0pt,
        ylabel = {Seconds},
        ylabel style = {scale=1.0},
        symbolic x coords={
        		A, B, C
        },
        xticklabels={
        		MK-Means, OCSVM, MMD
        },
	y tick label style={scale=0.9,
    		/pgf/number format/.cd,
   		fixed,
   		fixed zerofill,
    		precision=0},
        xtick = data,
        x tick label style={scale=0.9,
		},
        nodes near coords,
        every node near coord/.append style={rotate=90, scale=0.825,
        								   anchor=west, 
								   /pgf/number format/.cd,
								   fixed,
								   fixed zerofill,
								   precision=2},
        enlarge x limits=0.225,
        legend cell align=left,
        legend pos=north east,
        legend style={nodes={scale=0.85},
        },
]
\addplot [fill=blue,opacity=1.00]
coordinates {
(A, 1.0717)
(B, 0.0651)
(C, 0.6248)
};
\addlegendentry{\hbox{Min}}
\addplot [fill=green,opacity=1.00]
coordinates {
(A, 7.7096)
(B, 0.4646)
(C, 1.8544)
};
\addlegendentry{\hbox{Max}}
\addplot [fill=red,opacity=1.00]
coordinates {
(A, 3.1666)
(B, 0.1532)
(C, 1.0567)
};
\addlegendentry{\hbox{Mean}}
\addplot [fill=yellow,opacity=1.00]
coordinates {
(A, 1.3406)
(B, 0.0710)
(C, 0.3928)
};
\addlegendentry{\hbox{Standard deviation}}
\end{axis}
\end{tikzpicture}
    \caption{Time required to detect drift points per~$(F_{d}, F_{c})$ pair}
    \label{fig:timings}
\end{figure}
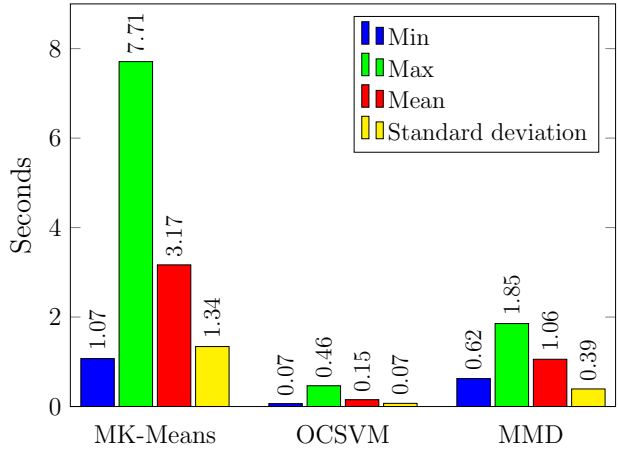

A comparison of the average time per~$(F_{d},F_{c})$ pair that is 
required for each of the three drift-detection 
techniques is given in the form of a bar graph 
in Figure~\ref{fig:timings}. By this measure, OCSVM requires the least time, by
a wide margin, with MK-Means requiring the most time, also by a wide margin.
The primary reason for these large discrepancies is that the OCSVM
model is only retrained when concept drift is detected, whereas 
MK-Means and MMD need to be updated for each block, 
regardless of whether drift is detected.
It is also worth noting that OCSVM has the largest number of hyperparameters
to test and, as discussed in Section~\ref{sec:mmdDiscussion}, 
this number could likely be significantly reduced, 
which would further increase its timing advantage.

According to a majority of the metrics considered, our OCSVM concept
drift detector outperforms the MK-Means and MMD approaches.
Thus, we conclude that OCSVM is the optimal choice from
among the three concept drift detection techniques analyzed in this chapter.

\section{Conclusion and Future Work}\label{chap:conclusion}

In this chapter, we considered the problem of concept drift detection when using machine learning models
for malware classification. We analyzed three distinct concept drift detection approaches, 
two of which use machine learning techniques, namely, Minibatch K-Means (MK-Means) 
and One-Class Support Vector Machine (OCSVM), 
and one of which uses Maximum Mean Discrepancy (MMD), a statistical analysis technique.
We compared drift-aware detection strategies based on each of these three techniques, and in
each case, we compared the results to a static scenario (no model retraining occurs), and
to periodic retraining (models are constantly retrained, regardless of whether concept 
drift has occurred).

We also performed an extensive investigation of the tradeoff between accuracy and training efficiency,
using Pareto Front analysis. In addition, we analyzed the time required to train each of the 
concept drift detectors.

Our OCSVM-based concept drift detection technique generally performed best, 
with respect to accuracy and training time, and
with respect to training efficiency, OCSVM was equivalent to MMD.
In terms of accuracy, efficiency, and training time,
MK-Means was generally
the least effective of the three drift detection techniques. 
Over the set of hyperparameters tested, 
we found that MMD was the most consistent, but for the best 
selections of hyperparameters, OCSVM performed somewhat 
better than MMD. 

It is interesting---and perhaps somewhat surprising---that a machine learning 
based approach (OCSVM) outperformed a statistical based approach (MMD).
One plausible explanation is that while MMD is more sensitive to
changes in the underlying statistics, OCSVM is better at detecting only those changes
that are likely to actually matter, from a machine learning model perspective. 
This is a topic that is worthy of further investigation.

An additional avenue for future research is refining 
the window size used for concept drift detection. While we only considered a fixed window
of size~50, the smallest practical window size would be preferred,
since we want to retrain our models as soon as malware has evolved. 
Additionally, a deeper analysis of the parameters of MK-Means and OCSVM 
could yield a better understanding of the underlying data distribution 
and thereby improve the accuracy and efficiency of these drift detectors.

Another promising topic for further exploration is the use of 
Long-Short Term Memory (LSTM) models as drift detectors.
LSTM models can account for temporal dependencies,
which might enable such models to dynamically learn the optimal window size
as part of the training process.
Finally, evaluating the performance of proposed drift detection techniques in 
real-time malware detection systems would provide valuable insights 
into their practical applicability and scalability.


\bibliographystyle{plain}
\bibliography{references}

\section*{Appendix}\label{app:a}

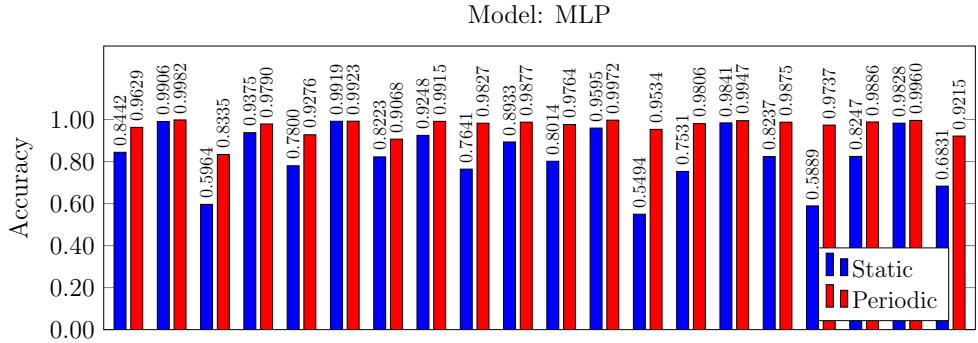
\begin{figure}[!htb]
    \centering
    \begin{tikzpicture}[scale=0.9, every node/.style={scale=1.0}]
\pgfkeys{/pgf/number format/.cd,1000 sep={}}
\begin{axis}[
        width  = 0.9*\textwidth,
        height = 5.75cm,
        ymin=0.0,ymax=1.35,
        ytick={0.0, 0.2, 0.4, 0.6, 0.8, 1.0},
        major x tick style = transparent,
        ybar=5*\pgflinewidth,
        bar width=5.0pt,
        ylabel = {Accuracy},
        ylabel style = {scale = 0.9},
        title = {Model: MLP},
        title style = {scale = 0.9},
        symbolic x coords={A, B, C, D, E, F, G, H, I, J, K, L, M, N, O, P, Q, R, S, T},
        xticklabels={(\Agent, \Airpush), (\Agent, \Boxer), (\Agent, \Malap), (\Agent, \SMSreg),
        			   (\Airpush, \Agent), (\Airpush, \Boxer), (\Airpush, \Malap), (\Airpush, \SMSreg),
			   (\Boxer, \Agent), (\Boxer, \Airpush), (\Boxer, \Malap), (\Boxer, \SMSreg),
			   (\Malap, \Agent), (\Malap, \Airpush), (\Malap, \Boxer), (\Malap, \SMSreg),
			   (\SMSreg, \Agent), (\SMSreg, \Airpush), (\SMSreg, \Boxer), (\SMSreg, \Malap),
        },
	y tick label style={scale=0.9,
    		/pgf/number format/.cd,
   		fixed,
   		fixed zerofill,
    		precision=2},
        xtick = data,
        x tick label style={scale=0.85,
        		rotate=60,
		anchor=north east,
		inner sep=0mm
		},
        nodes near coords,
        every node near coord/.append style={rotate=90, scale=0.65,
        								   anchor=west, 
								   /pgf/number format/.cd,
								   fixed,
								   fixed zerofill,
								   precision=4},
        enlarge x limits=0.03,
        legend cell align=left,
        legend pos=south east,
        xticklabels=\empty,
        legend style={nodes={scale=0.85}
        },
]
\addplot [fill=blue,opacity=1.00]
coordinates {
(A, 0.8442)
(B, 0.9906)
(C, 0.5964)
(D, 0.9375)
(E, 0.7800)
(F, 0.9919)
(G, 0.8223)
(H, 0.9248)
(I, 0.7641)
(J, 0.8933)
(K, 0.8014)
(L, 0.9595)
(M, 0.5494)
(N, 0.7531)
(O, 0.9841)
(P, 0.8237)
(Q, 0.5889)
(R, 0.8247)
(S, 0.9828)
(T, 0.6831)
};
\addlegendentry{Static}
\addplot [fill=red,opacity=1.00]
coordinates {
(A, 0.9629)
(B, 0.9982)
(C, 0.8335)
(D, 0.9790)
(E, 0.9276)
(F, 0.9923)
(G, 0.9068)
(H, 0.9915)
(I, 0.9827)
(J, 0.9877)
(K, 0.9764)
(L, 0.9972)
(M, 0.9534)
(N, 0.9806)
(O, 0.9947)
(P, 0.9875)
(Q, 0.9737)
(R, 0.9886)
(S, 0.9960)
(T, 0.9215)
};
\addlegendentry{Periodic}
\end{axis}
\end{tikzpicture}
    \caption{Static and periodic accuracies for MLP model}
    \label{fig:periodicStatic_MLP}
\end{figure}
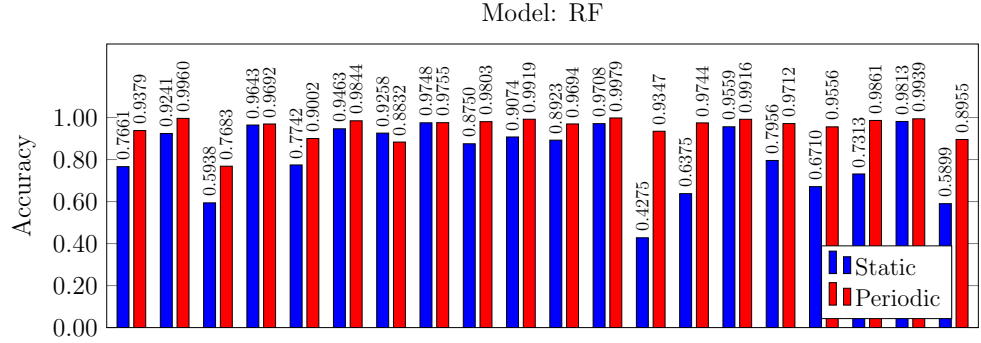
\begin{figure}[!htb]
    \centering
    \begin{tikzpicture}[scale=0.9, every node/.style={scale=1.0}]
\pgfkeys{/pgf/number format/.cd,1000 sep={}}
\begin{axis}[
        width  = 0.9*\textwidth,
        height = 5.75cm,
        ymin=0.0,ymax=1.35,
        ytick={0.0, 0.2, 0.4, 0.6, 0.8, 1.0},
        major x tick style = transparent,
        ybar=5*\pgflinewidth,
        bar width=5.0pt,
        ylabel = {Accuracy},
        ylabel style = {scale = 0.9},
        title = {Model: RF},
        title style = {scale = 0.9},
        symbolic x coords={A, B, C, D, E, F, G, H, I, J, K, L, M, N, O, P, Q, R, S, T},
        xticklabels={(\Agent, \Airpush), (\Agent, \Boxer), (\Agent, \Malap), (\Agent, \SMSreg),
        			   (\Airpush, \Agent), (\Airpush, \Boxer), (\Airpush, \Malap), (\Airpush, \SMSreg),
			   (\Boxer, \Agent), (\Boxer, \Airpush), (\Boxer, \Malap), (\Boxer, \SMSreg),
			   (\Malap, \Agent), (\Malap, \Airpush), (\Malap, \Boxer), (\Malap, \SMSreg),
			   (\SMSreg, \Agent), (\SMSreg, \Airpush), (\SMSreg, \Boxer), (\SMSreg, \Malap),
        },
	y tick label style={scale=0.9,
    		/pgf/number format/.cd,
   		fixed,
   		fixed zerofill,
    		precision=2},
        xtick = data,
        x tick label style={scale=0.85,
        		rotate=60,
		anchor=north east,
		inner sep=0mm
		},
        nodes near coords,
        every node near coord/.append style={rotate=90, scale=0.65,
        								   anchor=west, 
								   /pgf/number format/.cd,
								   fixed,
								   fixed zerofill,
								   precision=4},
        enlarge x limits=0.03,
        legend cell align=left,
        legend pos=south east,
        xticklabels=\empty,
        legend style={nodes={scale=0.85}
        },
]
\addplot [fill=blue,opacity=1.00]
coordinates {
(A, 0.7661)
(B, 0.9241)
(C, 0.5938)
(D, 0.9643)
(E, 0.7742)
(F, 0.9463)
(G, 0.9258)
(H, 0.9748)
(I, 0.8750)
(J, 0.9074)
(K, 0.8923)
(L, 0.9708)
(M, 0.4275)
(N, 0.6375)
(O, 0.9559)
(P, 0.7956)
(Q, 0.6710)
(R, 0.7313)
(S, 0.9813)
(T, 0.5899)
};
\addlegendentry{Static}
\addplot [fill=red,opacity=1.00]
coordinates {
(A, 0.9379)
(B, 0.9960)
(C, 0.7683)
(D, 0.9692)
(E, 0.9002)
(F, 0.9844)
(G, 0.8832)
(H, 0.9755)
(I, 0.9803)
(J, 0.9919)
(K, 0.9694)
(L, 0.9979)
(M, 0.9347)
(N, 0.9744)
(O, 0.9916)
(P, 0.9712)
(Q, 0.9556)
(R, 0.9861)
(S, 0.9939)
(T, 0.8955)
};
\addlegendentry{Periodic}
\end{axis}
\end{tikzpicture}
    \caption{Static and periodic accuracies for RF model}
    \label{fig:periodicStatic_RF}
\end{figure}
\begin{figure}[!htb]
    \centering
    \begin{tikzpicture}[scale=0.9, every node/.style={scale=1.0}]
\pgfkeys{/pgf/number format/.cd,1000 sep={}}
\begin{axis}[
        width  = 0.9*\textwidth,
        height = 5.75cm,
        ymin=0.0,ymax=1.35,
        ytick={0.0, 0.2, 0.4, 0.6, 0.8, 1.0},
        major x tick style = transparent,
        ybar=5*\pgflinewidth,
        bar width=5.0pt,
        ylabel = {Accuracy},
        ylabel style = {scale = 0.9},
        title = {Model: XGB},
        title style = {scale = 0.9},
        symbolic x coords={A, B, C, D, E, F, G, H, I, J, K, L, M, N, O, P, Q, R, S, T},
        xticklabels={(\Agent, \Airpush), (\Agent, \Boxer), (\Agent, \Malap), (\Agent, \SMSreg),
        			   (\Airpush, \Agent), (\Airpush, \Boxer), (\Airpush, \Malap), (\Airpush, \SMSreg),
			   (\Boxer, \Agent), (\Boxer, \Airpush), (\Boxer, \Malap), (\Boxer, \SMSreg),
			   (\Malap, \Agent), (\Malap, \Airpush), (\Malap, \Boxer), (\Malap, \SMSreg),
			   (\SMSreg, \Agent), (\SMSreg, \Airpush), (\SMSreg, \Boxer), (\SMSreg, \Malap),
        },
	y tick label style={scale=0.9,
    		/pgf/number format/.cd,
   		fixed,
   		fixed zerofill,
    		precision=2},
        xtick = data,
        x tick label style={scale=0.85,
        		rotate=60,
		anchor=north east,
		inner sep=0mm
		},
        nodes near coords,
        every node near coord/.append style={rotate=90, scale=0.65,
        								   anchor=west, 
								   /pgf/number format/.cd,
								   fixed,
								   fixed zerofill,
								   precision=4},
        enlarge x limits=0.03,
        legend cell align=left,
        legend pos=south east,
        legend style={nodes={scale=0.85}
        },
]
\addplot [fill=blue,opacity=1.00]
coordinates {
(A, 0.7362)
(B, 0.9737)
(C, 0.5746)
(D, 0.8246)
(E, 0.6426)
(F, 0.9984)
(G, 0.6863)
(H, 0.8256)
(I, 0.6518)
(J, 0.7077)
(K, 0.8062)
(L, 0.9430)
(M, 0.5425)
(N, 0.5172)
(O, 0.8359)
(P, 0.8566)
(Q, 0.6306)
(R, 0.8982)
(S, 0.8177)
(T, 0.6328)
};
\addlegendentry{Static}
\addplot [fill=red,opacity=1.00]
coordinates {
(A, 0.9388)
(B, 0.9897)
(C, 0.7879)
(D, 0.9420)
(E, 0.7926)
(F, 0.9795)
(G, 0.9655)
(H, 0.9684)
(I, 0.8754)
(J, 0.9634)
(K, 0.9215)
(L, 0.9771)
(M, 0.9053)
(N, 0.9634)
(O, 0.9803)
(P, 0.9694)
(Q, 0.9197)
(R, 0.9702)
(S, 0.9803)
(T, 0.8778)
};
\addlegendentry{Periodic}
\end{axis}
\end{tikzpicture} 
    \caption{Static and periodic accuracies for XGB model}
    \label{fig:periodicStatic_XGB}
\end{figure}
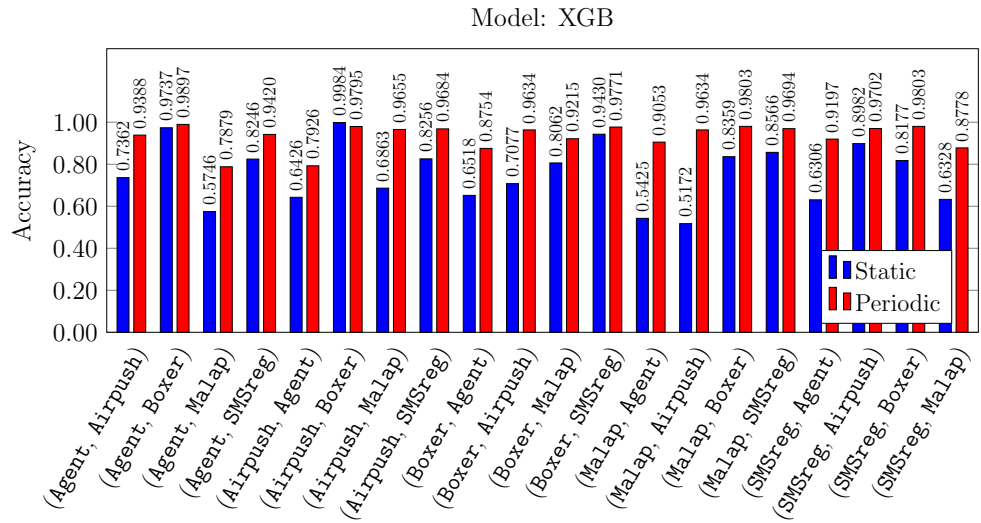
\begin{figure}[!htb]
    \centering
    \includegraphics[width=0.8\textwidth]{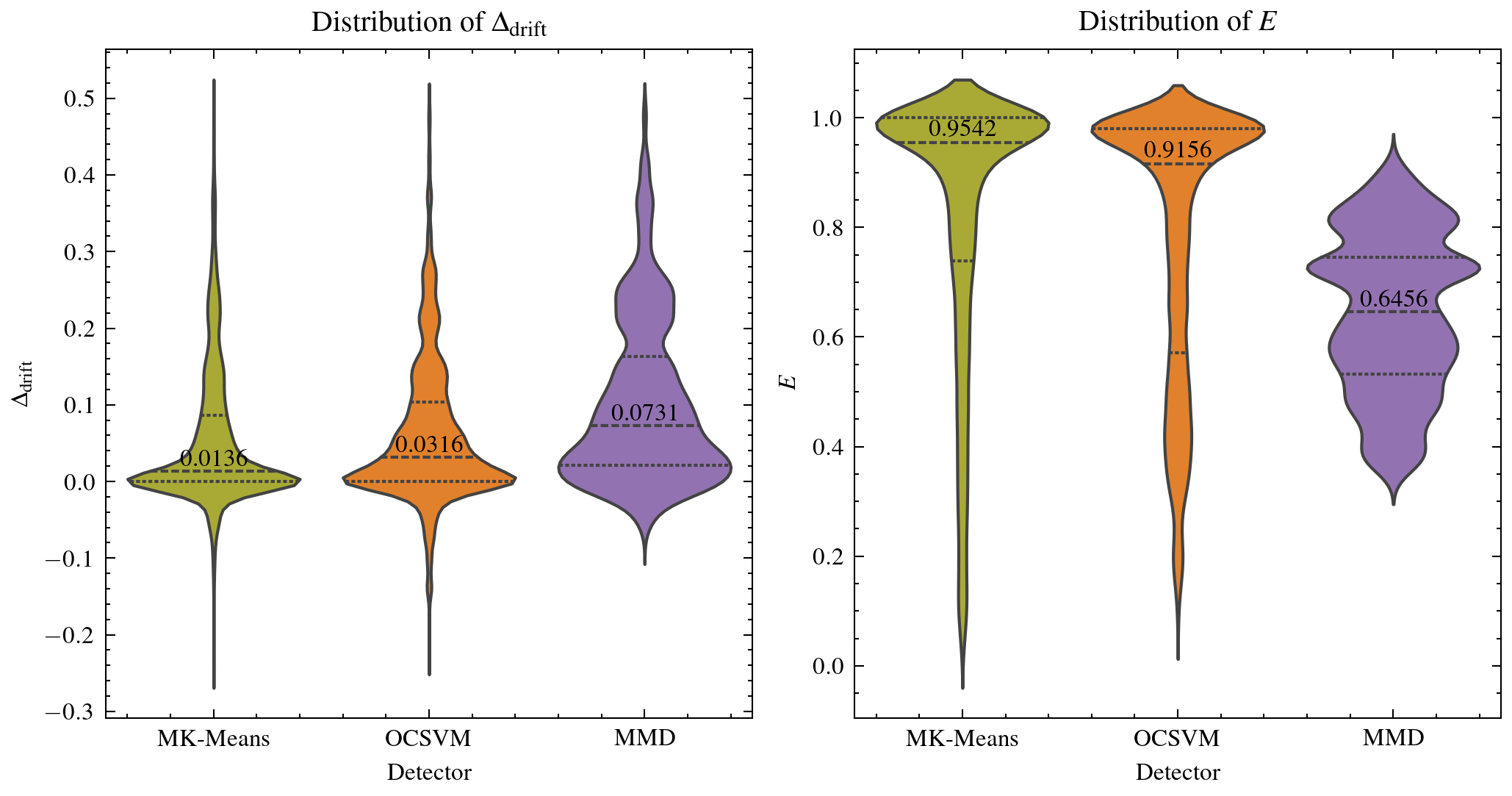}
    \caption{Violin plots for accuracy and efficiency over all hyperparameters}
    \label{fig:violinAll}
\end{figure}
\begin{figure}[!htb]
    \centering
    \includegraphics[width=0.8\textwidth]{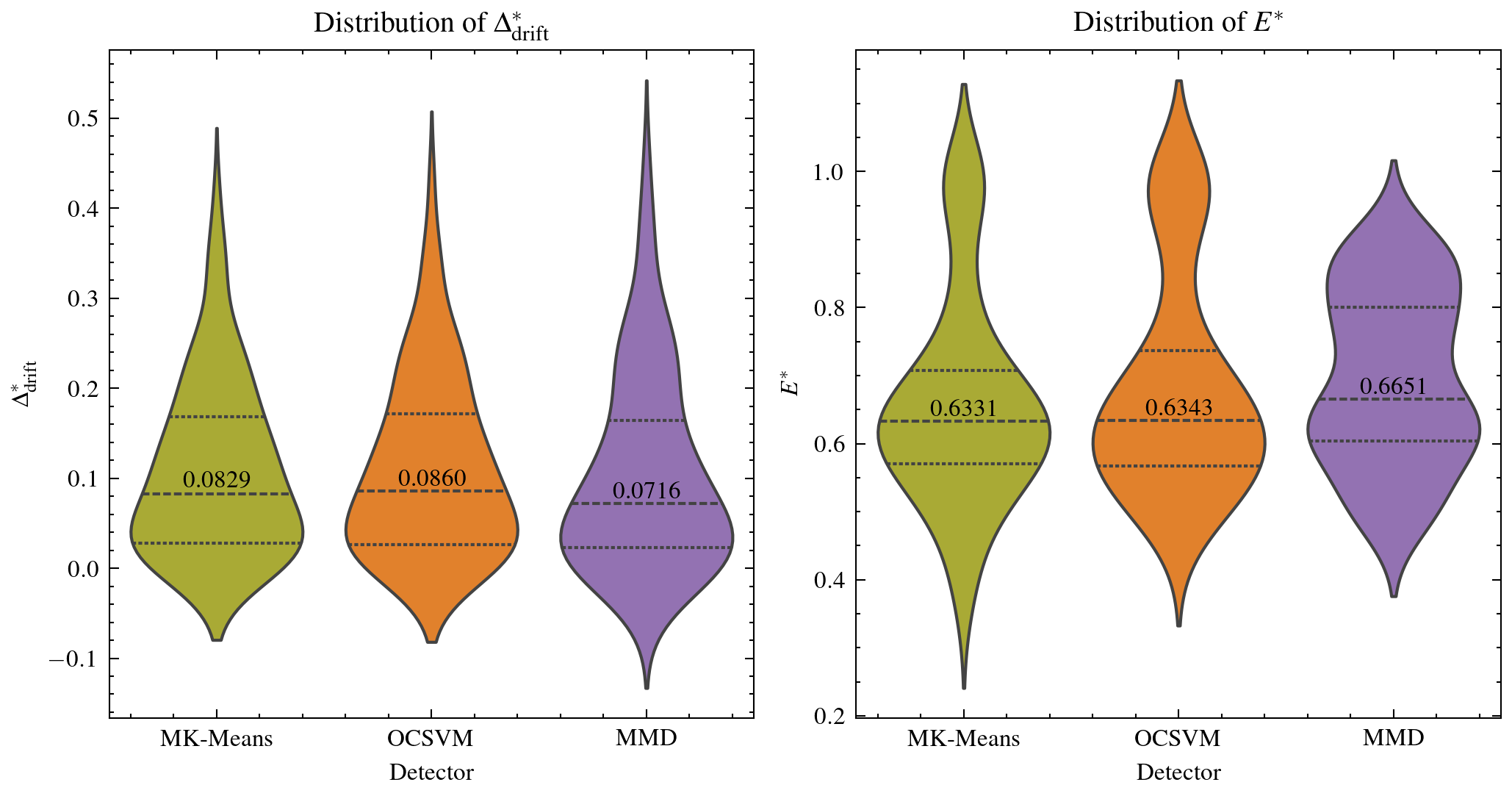}
    \caption{Violin plots for accuracy and efficiency on Pareto Front}
    \label{fig:violinPareto}
\end{figure}

In this Appendix, we provide additional relevant graphs.
Figures~\ref{fig:periodicStatic_MLP} through~\ref{fig:periodicStatic_XGB} 
provide bar graphs for each
learning model considered, where static and periodic accuracies are given
for each combination of malware families~$(F_{d},F_{c})$.
Figure~\ref{fig:violinAll} contains violin plots related to
accuracy and efficiency over all hyperparameters tested,
while Figure~\ref{fig:violinPareto} has the analogous
plots restricted to the corresponding Pareto Front.

\begin{figure}[!htb]
    \centering
    \begin{tikzpicture}[scale=0.9, every node/.style={scale=1.0}]
\pgfkeys{/pgf/number format/.cd,1000 sep={}}
\begin{axis}[
        width  = 0.9*\textwidth,
        height = 5.75cm,
        ymin=0.0,ymax=1.35,
        ytick={0.0, 0.2, 0.4, 0.6, 0.8, 1.0},
        major x tick style = transparent,
        ybar=5*\pgflinewidth,
        bar width=5.0pt,
        ylabel = {Accuracy},
        ylabel style = {scale = 0.9},
        title = {Model: SVM},
        title style = {scale = 0.9},
        symbolic x coords={A, B, C, D, E, F, G, H, I, J, K, L, M, N, O, P, Q, R, S, T},
        xticklabels={(\Agent, \Airpush), (\Agent, \Boxer), (\Agent, \Malap), (\Agent, \SMSreg),
        			   (\Airpush, \Agent), (\Airpush, \Boxer), (\Airpush, \Malap), (\Airpush, \SMSreg),
			   (\Boxer, \Agent), (\Boxer, \Airpush), (\Boxer, \Malap), (\Boxer, \SMSreg),
			   (\Malap, \Agent), (\Malap, \Airpush), (\Malap, \Boxer), (\Malap, \SMSreg),
			   (\SMSreg, \Agent), (\SMSreg, \Airpush), (\SMSreg, \Boxer), (\SMSreg, \Malap),
        },
	y tick label style={scale=0.9,
    		/pgf/number format/.cd,
   		fixed,
   		fixed zerofill,
    		precision=2},
        xtick = data,
        x tick label style={scale=0.85,
        		rotate=60,
		anchor=north east,
		inner sep=0mm
		},
        nodes near coords,
        every node near coord/.append style={rotate=90, scale=0.65,
        								   anchor=west, 
								   /pgf/number format/.cd,
								   fixed,
								   fixed zerofill,
								   precision=4},
        enlarge x limits=0.03,
        legend cell align=left,
        legend pos=south east,
        xticklabels=\empty,
        legend style={nodes={scale=0.85}
        },
]
\addplot [fill=blue,opacity=1.00]
coordinates {
(A, 0.7469)
(B, 0.9335)
(C, 0.5835)
(D, 0.8920)
(E, 0.8668)
(F, 0.9450)
(G, 0.9402)
(H, 0.8860)
(I, 0.8683)
(J, 0.9345)
(K, 0.8384)
(L, 0.9606)
(M, 0.6894)
(N, 0.6478)
(O, 0.9547)
(P, 0.8578)
(Q, 0.7851)
(R, 0.8111)
(S, 0.9881)
(T, 0.7015)
};
\addlegendentry{Static}
\addplot [fill=red,opacity=1.00]
coordinates {
(A, 0.8942)
(B, 0.9621)
(C, 0.8848)
(D, 0.9192)
(E, 0.9174)
(F, 0.9835)
(G, 0.9331)
(H, 0.9155)
(I, 0.9856)
(J, 0.9835)
(K, 0.9835)
(L, 0.9852)
(M, 0.9281)
(N, 0.9478)
(O, 0.9891)
(P, 0.9487)
(Q, 0.9540)
(R, 0.9558)
(S, 0.9960)
(T, 0.9326)
};
\addlegendentry{Periodic}
\end{axis}
\end{tikzpicture}
    \caption{Static and periodic accuracies for SVM model}
    \label{fig:periodicStatic_SVM}
\end{figure}

\end{document}